\documentclass[journal]{IEEEtran}

\bstctlcite{bstctl:etal, bstctl:nodash, bstctl:simpurl}

\usepackage{booktabs}
\usepackage{caption}
\usepackage{gensymb}
\usepackage{graphicx}
\usepackage{multirow}
\usepackage{url}
\usepackage{xspace}
\usepackage{xfrac}
\usepackage[colorlinks,pagebackref=false,citecolor=blue,bookmarks=false,hypertexnames=true]{hyperref}

\begin{document}

\bstctlcite{IEEEexample:BSTcontrol}

\title{An Online Learning System for Wireless Charging Alignment using Surround-view Fisheye Cameras}

\author{Ashok~Dahal,
        Varun~Ravi~Kumar,
        Senthil~Yogamani,
        and~Ciar\' {a}n Eising,~\IEEEmembership{Member,~IEEE
        }%

\thanks{A. Dahal is with Valeo North America, Troy, MI, USA. E-mail: ashok.dahal@valeo.com.}%
\thanks{V. Ravi Kumar is with Valeo Driving Assistance Research, Kronach, Germany. E-mail: varun.ravi-kumar@valeo.com.}%
\thanks{S. Yogamani is with the Valeo Vision Systems, Tuam, Co. Galway, Ireland. E-mail: senthil.yogamani@valeo.com.}%
\thanks{C. Eising is with the Department of Electronic and Computer Engineering, University of Limerick, Ireland. E-mail: ciaran.eising@ul.ie.}
\thanks{Corresponding author: C. Eising (email: ciaran.eising@ul.ie)}
}

\markboth{IEEE TRANSACTIONS ON INTELLIGENT TRANSPORTATION SYSTEMS,~Vol.~XX, No.~X, XXX~202X}%
{Dahal \MakeLowercase{\textit{et al.}}: An Online Learning System for Wireless Charging Alignment using Surround-view Fisheye Cameras}


\maketitle
\begin{abstract}
Electric Vehicles are increasingly common, with inductive chargepads being considered a convenient and efficient means of charging electric vehicles. However, drivers are typically poor at aligning the vehicle to the necessary accuracy for efficient inductive charging, making the automated alignment of the two charging plates desirable. In parallel to the electrification of the vehicular fleet, automated parking systems that make use of surround-view camera systems are becoming increasingly popular. In this work, we propose a system based on the surround-view camera architecture to detect, localize, and automatically align the vehicle with the inductive chargepad. The visual design of the chargepads is not standardized and not necessarily known beforehand. Therefore, a system that relies on offline training will fail in some situations. Thus, we propose a self-supervised online learning method that leverages the driver's actions when manually aligning the vehicle with the chargepad and combine it with weak supervision from semantic segmentation and depth to learn a classifier to auto-annotate the chargepad in the video for further training. In this way, when faced with a previously unseen chargepad, the driver needs only manually align the vehicle a single time. As the chargepad is flat on the ground, it is not easy to detect it from a distance. Thus, we propose using a Visual SLAM pipeline to learn landmarks relative to the chargepad to enable alignment from a greater range. We demonstrate the working system on an automated vehicle as illustrated in the video \url{https://youtu.be/_cLCmkW4UYo}. To encourage further research, we will share a chargepad dataset used in this work\footnote{An initial version of the dataset is shared \href{https://drive.google.com/drive/folders/1KeLFIqOnhU2CGsD0vbiN9UqKmBSyHERd}{here}.}. 

\end{abstract}
\section{Introduction}

\IEEEPARstart{W}{ith} predicted growth in the electric vehicle market over the next decade, with a market penetration of up to 42.5\% by 2035 expected \cite{RIETMANN2020121038}, wireless charging is seen as a key enabling technology and is now the subject of standardization \cite{sae2020}. Inductive charging is market-ready \cite{HE20131} but suffers due to driver behavior. Inductive charging efficiency is highly related to the charging coils' position relative to one another, with a tolerance of approximately $\pm 10$cm from the center points. In contrast, studies have shown that human parking accuracy is between 20 and 120cm longitudinally and 20 to 60cm laterally, with only 5\% of vehicles parked within the tolerances for optimal efficiency \cite{BIRRELL2015721}. \par

\begin{figure}[t]
\centering
\captionsetup{singlelinecheck=false, font=small, belowskip=-4pt}
\includegraphics[width=0.35\textwidth]{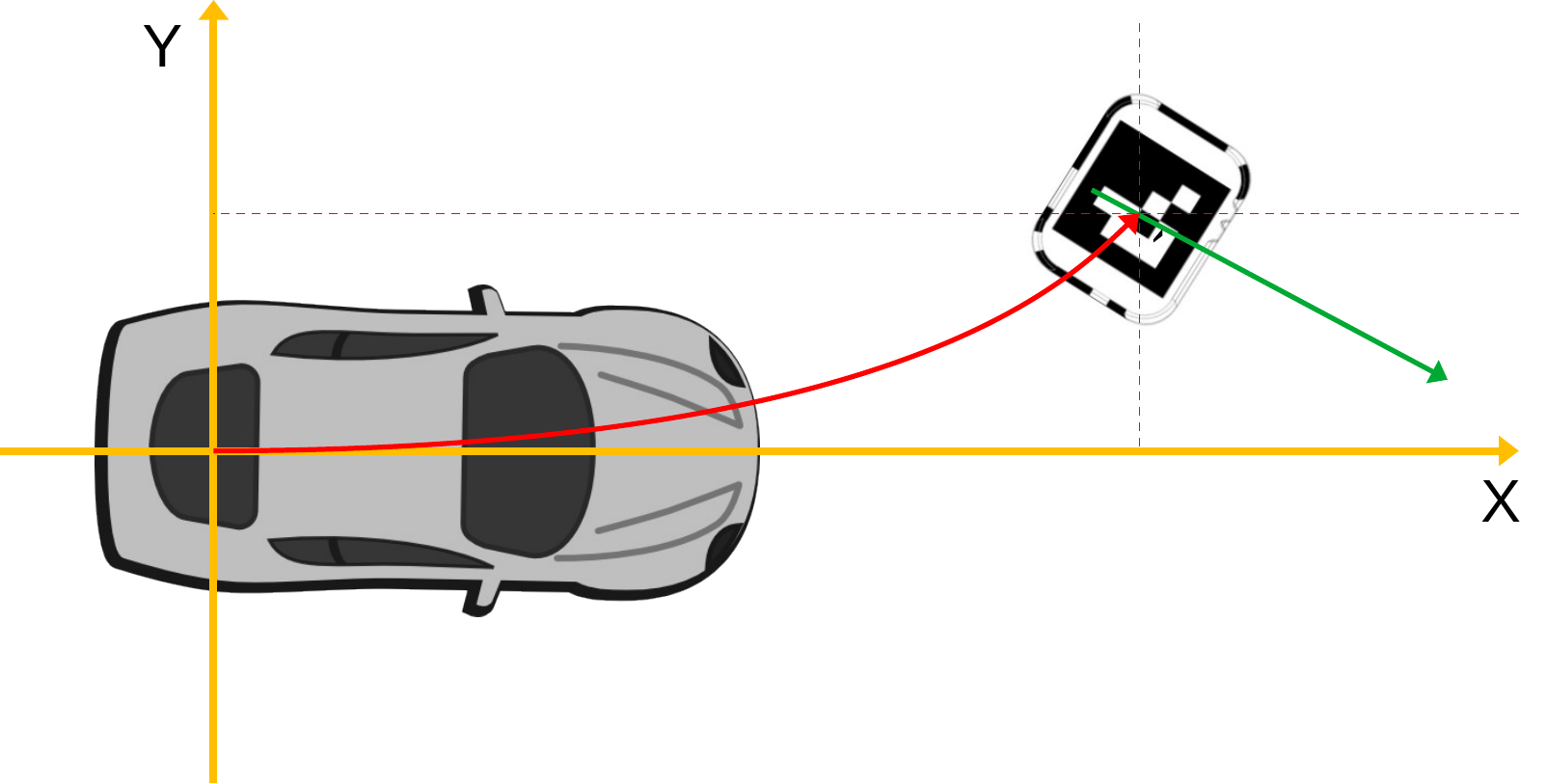}
\llap{\scalebox{-1}[1]{
      \includegraphics[width=0.1\textwidth]{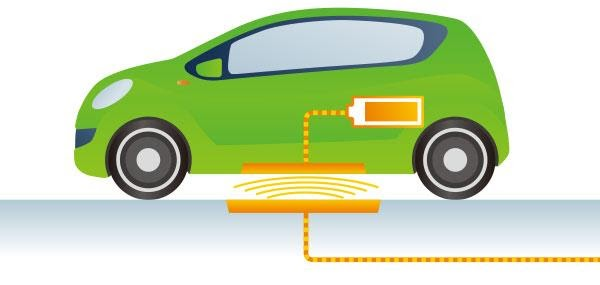}%
    }}
\caption{\textbf{Illustration of Chargepad Parking, including detection, localization, and alignment with automated driving.} The positional alignment (indicated by the red line) and the angular alignment (green axis) are both important for optimal charging.}
\label{fig:CpadParking}
\vspace{-0.4cm}
\end{figure}

Given that manual alignment of the charging coil is generally not accurate enough for optimal charging performance, an apparent problem exists that is yet to be completely resolved: that of the automation of the alignment of a vehicle with the chargepad (Figure \ref{fig:CpadParking}). This automated alignment is required to achieve high charging efficiency and thus increase uptake of wireless charging for electric vehicles. It is the objective of this paper to propose a computer vision system based on commonly deployed surround-view cameras that can be used to detect the chargepad and thus support the alignment of the charging coils. Notably, we aim to address the problem of previously unseen chargepad designs using online learning.\par
Several solutions have been proposed to solve the problem of automated chargepad alignment. In \cite{Ni2015}, a system whereby additional hardware in the form of wireless sensors on the coils is proposed, with at least six sensors (three per coil) required. In \cite{Babu2019}, a set of Inductor-capacitor (\textit{LC}) sensors is proposed to be installed on the installed charging coil, enabling the detection of the direction that the electric vehicle must be driven for optimal alignment. Tiemann et al. \cite{Tiemann2016} propose an ultra-wideband wireless alignment system based on IEEE 802.15.4a, and \cite{Senju2019} propose a method of position detection based on the analysis of voltage differences in so-called search coils. These methods require specific hardware to be installed on the vehicle and/or the chargepad to enable the alignment. It is also not clear how close the vehicle must be to the chargepad for these solutions to work. In contrast, our proposal is based on using a set of surround-view cameras that are standard on many vehicles, with a specific aim of increasing the range at which the maneuver automation can start. \par

\begin{figure}[t]
\centering
\captionsetup{singlelinecheck=false, font=small, belowskip=-4pt}
\includegraphics[width=\columnwidth,page=2]{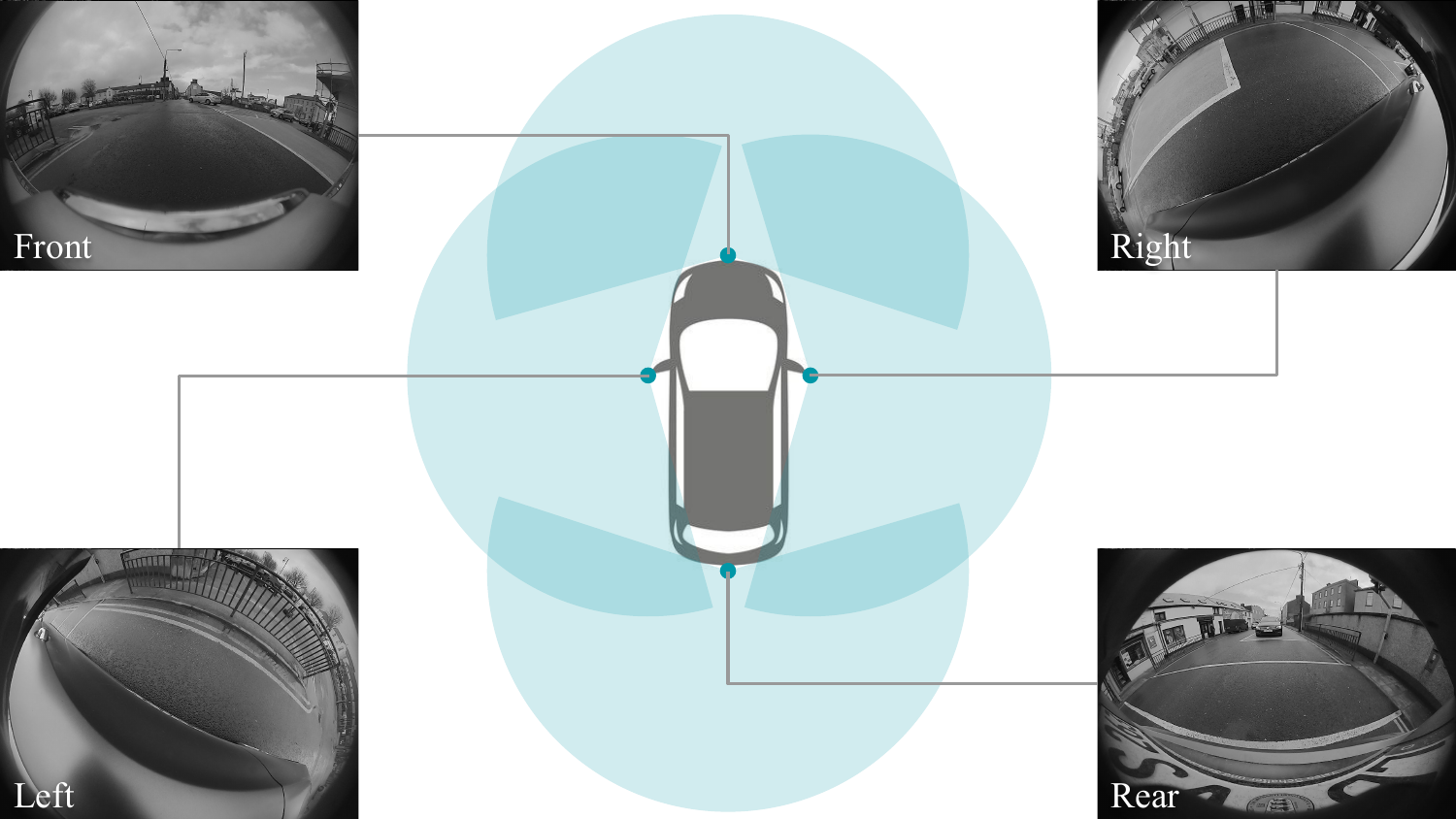}
\caption{\textbf{Illustration of a typical automotive surround-view system} consisting of four fisheye cameras located at the front, rear and on each wing-mirror.}
\label{fig:surroundview}
\vspace{-0.4cm}
\end{figure}

In parallel to the accelerated development of electric vehicles and wireless charging stations, wide field-of-view or fisheye cameras are becoming commonplace on vehicles, particularly for automated parking systems \cite{HEIMBERGER201788}, a problem that is highly related to the problem of wireless charging station alignment. Of key importance is that the fisheye cameras provide video of the scene right up to the vehicle body, providing important near-field sensing. In particular, fisheye cameras are commonly deployed on vehicles in a surround-view configuration \cite{HEIMBERGER201788}, as stylized in Figure \ref{fig:surroundview}, with one camera at both the front and rear of the vehicle and on each of the wing-mirrors. Such camera networks have traditionally been deployed for viewing applications (e.g., \cite{Liu2008} and many commercial surround-view products). However, surround-view cameras are increasingly focused on near field sensing, which are typically used for low vehicle speed applications such as parking or traffic jam assistance functions~\cite{Eising2021}. There is relatively less work on using convolutional neural networks for fisheye cameras. Recently, it has been explored for various tasks such as object detection \cite{rashed2021generalized, sekkat2022synwoodscape}, soiling detection \cite{uricar2021let, das2020tiledsoilingnet}, motion segmentation \cite{yahiaoui2019fisheyemodnet}, road edge detection \cite{dahal2021roadedgenet}, weather classification \cite{dhananjaya2021weather}, depth prediction \cite{kumar2021svdistnet, kumar2020fisheyedistancenet, kumar2021fisheyedistancenet++, kumar2020unrectdepthnet}, SLAM \cite{tripathi2020trained, gallagher2021hybrid, cheke2022fisheyepixpro} and in general for multi-task outputs \cite{kumar2021omnidet, klingner2022detecting, sobh2021adversarial, kia_2021}. 

There has been little previous work in chargepad alignment using computer vision approaches. A stereo vision-based approach has been proposed in the past \cite{Liu2017}. However, stereo vision systems suffer from field-of-view issues. They are typically unable to cover the vehicle's near-field area. When the chargepad moves out of the camera's view, one must solely rely on the vehicle odometry, which can lead to drift in the positional alignment. Additionally, the approach described in \cite{Liu2017} requires a specific design element of the chargepad, namely an LED at the center, which cannot be guaranteed for all chargepad designs.\par
\begin{figure*}[t]
\captionsetup{singlelinecheck=false, font=small, belowskip=-4pt}
\centering
\includegraphics[width=0.25\textwidth]{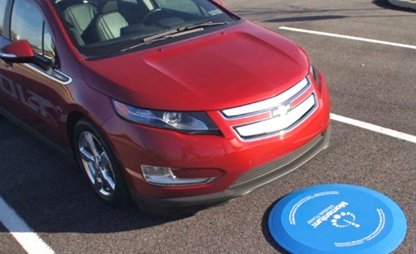}
\includegraphics[width=0.25\textwidth]{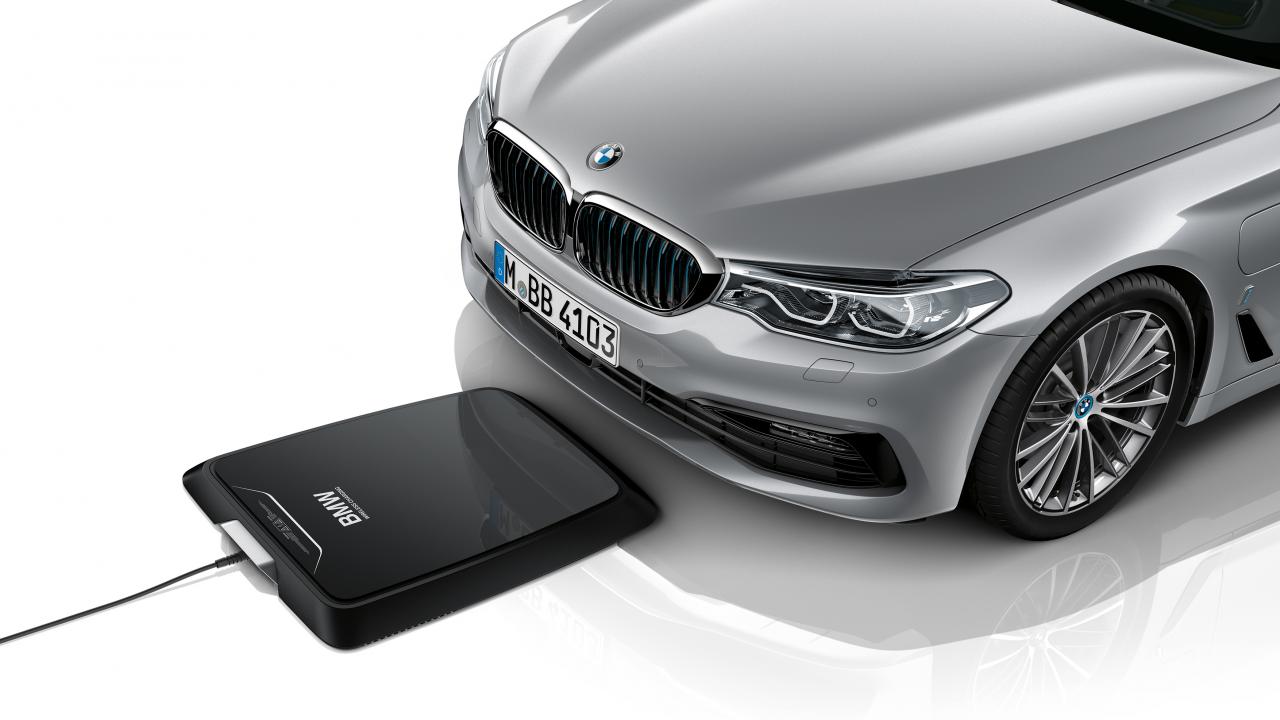}
\includegraphics[width=0.25\textwidth]{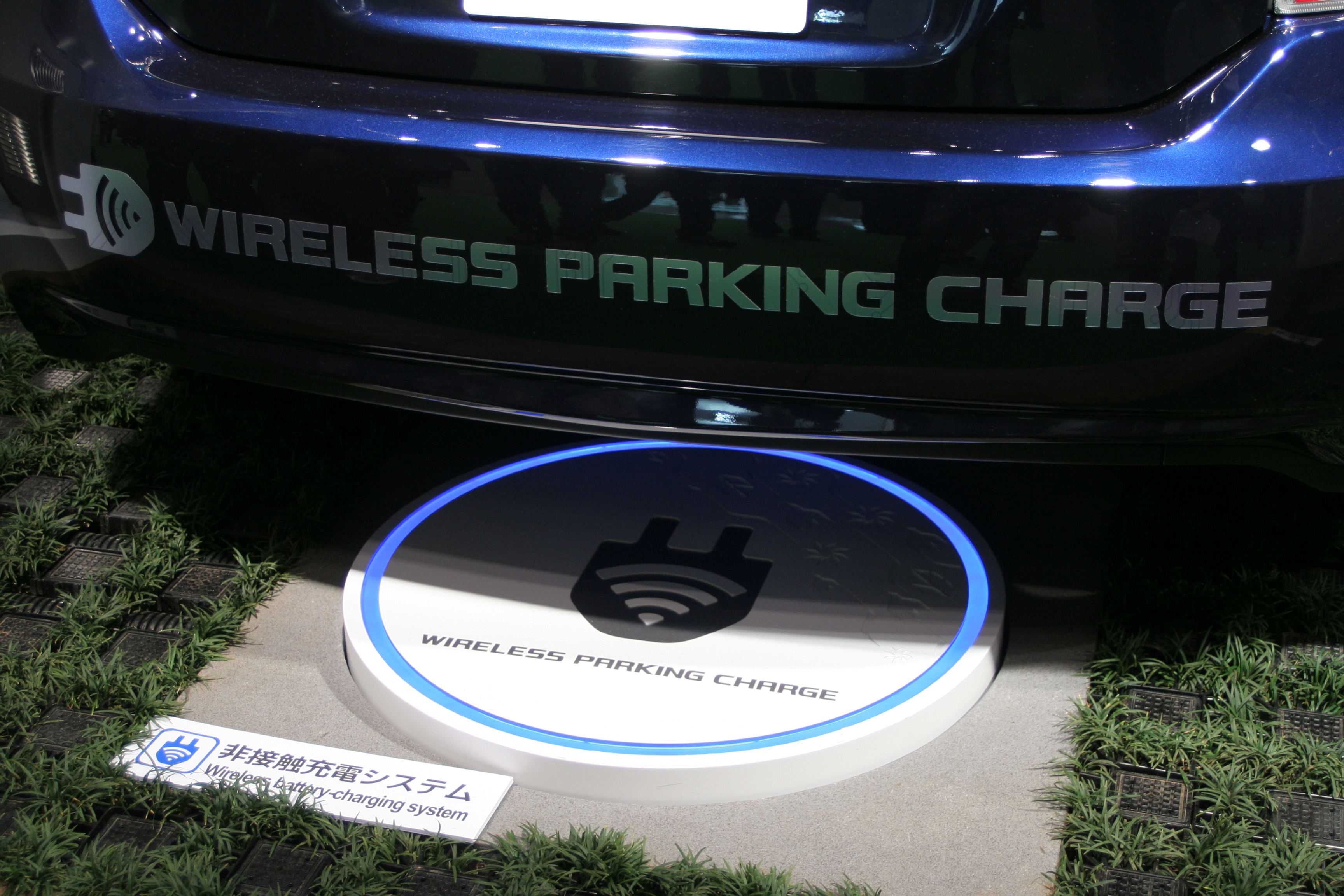}
\caption{\textbf{Various different commercial chargepads.} Images are taken from 
\href{https://www.hybridcars.com/momentum-dynamics-says-wireless-electric-car-charging-is-ready-now/}{[link1]}
\href{https://www.motorauthority.com/news/1124518_bmw-s-wireless-charging-system-reaches-us}{[link2]}
\href{https://commons.wikimedia.org/wiki/File:Electric_car_wireless_parking_charge_closeup.jpg}{[link3]}.
}
\label{fig:diffCpads}
\vspace{-0.4cm}
\end{figure*}

A problem with any visual approach to detecting the chargepad is that there are significantly different designs. An example of just a few is shown in Figure \ref{fig:diffCpads}. As electric vehicles gain momentum, different Original Equipment Manufacturers (OEMs) develop their version of electric vehicles and chargepad designs. As this is still in an early phase, the standardization of the chargepads has not been done yet. It may not happen due to aesthetic driving factors. In this paper, therefore, we propose an online self-supervised training method for solving the problem of the detection of unseen chargepads. An initial network is trained using a set of pre-defined chargepad types. If this fails for any chargepad alignment instance, the driver will need to manually align the vehicle with the chargepad. However, we propose a method to auto-annotate the chargepad by reversing the vehicle odometry to extract the location of the chargepad in recorded frames of video. We propose to solve accuracy issues with this approach by applying a refinement step to use semantic road segmentation and depth estimation to improve the localization of the chargepad in the video frames.\par
The main contribution of this paper is the proposed computer vision architecture, based on several algorithmic methods, that solves the problem of the alignment of a vehicle charging coil with the coil of the chargepad. We propose to detect the chargepad as part of multi-task neural network architecture. The alignment issues to previously unseen chargepad designs is solved by deploying a self-supervised online learning system, using an auto-annotation procedure based on vehicle odometry and semantic segmentation. The range of the alignment maneuver is increased through the fusion of the direct detection of the chargepads with Visual Simultaneous Localization and Mapping (SLAM). \par
The remainder of this paper is structured as follows. In Section \ref{sec:method}, we discuss the proposed method, describing the multi-task learning (MTL) perception stack, the Visual SLAM that we ultimately use for extending the range of the alignment algorithm, the online self-supervised learning method, and the overall system architecture proposal. We also describe a fiducial marker-based chargepad design. We then present a set of results that support our argument that this is an effective system for chargepad sensing and alignment in Section \ref{sec:results}.\par

\section{Proposed Methods and Architecture} \label{sec:method}

In this section, we describe the different methods that contribute to the overall architecture. We describe an ArUco pattern-based chargepad design, which we use in the test dataset, and which provides a baseline algorithm for our proposed approach. Then we describe the three-task perception stack that outputs object detection, segmentation, and distance that are used in the online auto-annotation later. We discuss how we approach online learning for the chargepad. We discuss Visual SLAM approaches that can be used to extend the range of the alignment algorithm beyond the limited range of direct detection of the chargepad. Finally, we introduce our overall architecture, whereby we combine all these elements.

\subsection{Fiducial Marker Chargepad Design}

Fiducial marker-based camera pose estimation or object detection and tracking are widely used in computer vision applications \cite{Kalaitzakis2021}. The configurable ArUco markers \cite{aruco} can be used as a fixed object for detection and tracking purposes in challenging scenes where various objects are present simultaneously. They have good contrast with both bright and dark regions. Additionally, they do not have rotational symmetry, and as such, orientation information can be extracted. As camera sensors are widely used, their cost is coming down. At the same time, quality such as image resolution is going up. The increase of computation time with the increase of image resolution can be well handled using simple ArUco based patterns in object detection and tracking \cite{speed-aruco}. 

\begin{figure}[ht]
\centering
\captionsetup{singlelinecheck=false, font=small, belowskip=-4pt}
\includegraphics[width=0.8\columnwidth]{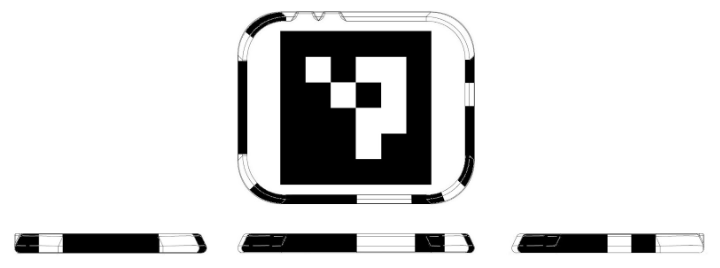}
\caption{\textbf{ArUco patterns on a charging station.}}
\label{fig:ArUco}
\vspace{-0.4cm}
\end{figure}

In this work, the architecture we design is not limited to ArUco patterns. In fact, having such patterns on the chargepad is non-desirable in OEM production, as they would not be considered aesthetically pleasing. However, using ArUco patterns is useful, as it gives a ground truth algorithm on which to base our development on. Therefore, our test data consists of chargepads with an ArUco pattern. The design of the chargepad pattern is demonstrated in Figure \ref{fig:ArUco}.

\subsection{Near-field Perception Stack through Multi-Task Learning (MTL)}

\begin{figure}[t]
\centering
\captionsetup{singlelinecheck=false, font=small, belowskip=-4pt}
\includegraphics[width=\columnwidth]{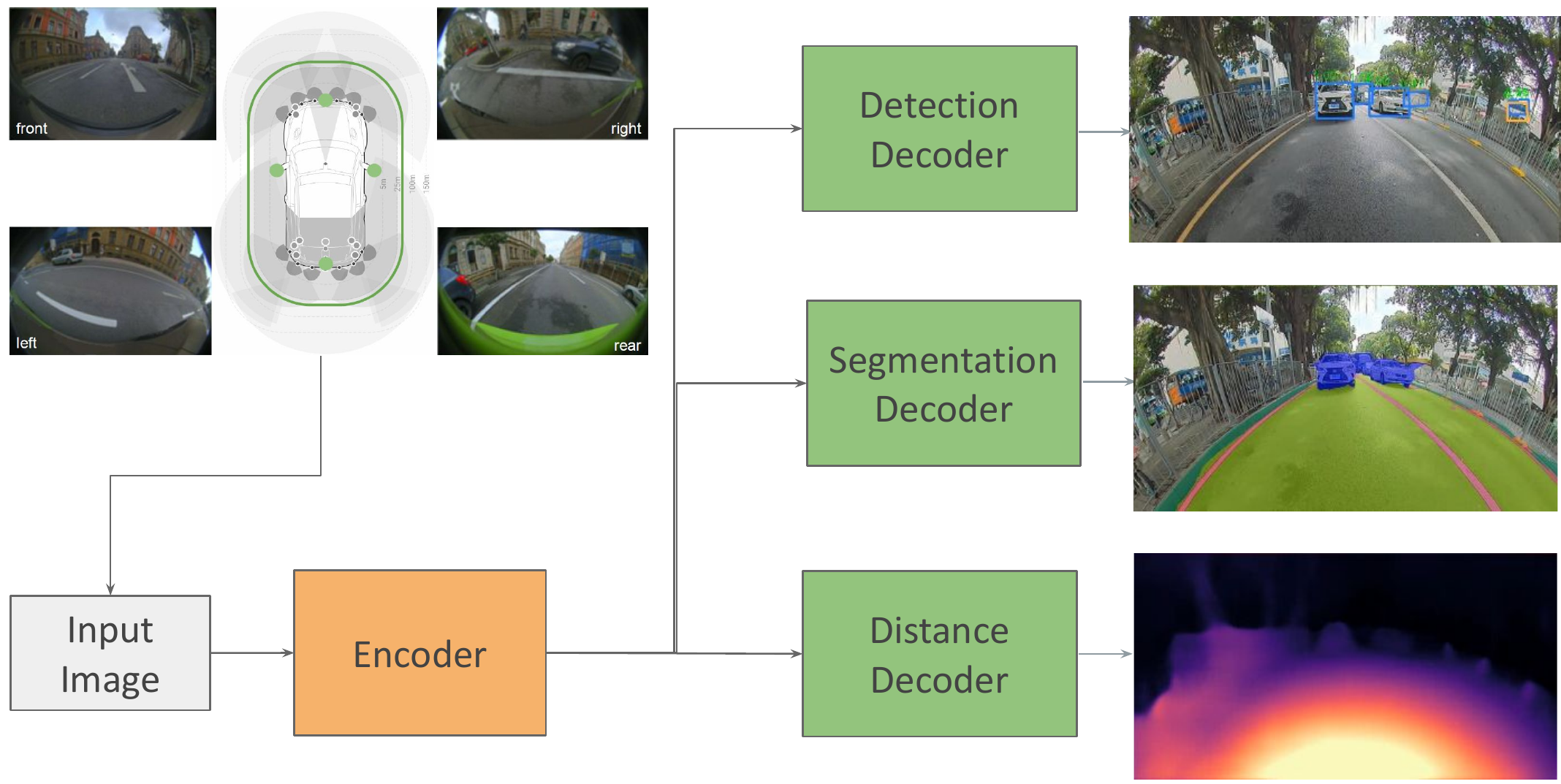}
\caption{\textbf{Three task perception stack.}}
\label{fig:perception}
\vspace{-0.2cm}
\end{figure}

We derive the baseline MTL model from our recent work OmniDet~\cite{kumar2021omnidet}, a six-task complete perception model for surround-view fisheye cameras. We focus on the three relevant perception tasks: object detection, semantic segmentation, and depth estimation. While we provide a short overview of the baseline model here, the reader is referred to \cite{kumar2021omnidet} for more complete details.
A high-level architecture of the model is shown in Figure \ref{fig:perception}. It comprises a shared ResNet18 \cite{He2016ResNet18} encoder and three decoders for each task. 2D bounding box detection task has the six important objects, namely \textit{pedestrians}, \textit{vehicles}, \textit{riders}, \textit{traffic sign}, \textit{traffic lights}. We add a new \textit{chargepads} class to the baseline.
Segmentation task has \textit{vehicles}, \textit{pedestrians}, \textit{cyclists}, \textit{road}, \textit{lanes}, and \textit{curbs} categories. The depth task provides scale-aware distance in 3D space. The model is trained jointly using the public WoodScape~\cite{yogamani2019woodscape} dataset comprising 8k samples and evaluated on 2k samples, augmented by our chargepad data.\par

The chargepad is first detected by the object detection part of our three task network as described in Figure \ref{fig:perception}. Like any other vision-based approach, our chargepad detection also suffers from issues like varying lighting and weather conditions, shadows \cite{Yahiaoui2019}, and the wear and tear of the pattern itself. So consistent detection of chargepad in each frame is not guaranteed. So, it is beneficial to track the previously detected bounding boxes to predict the relative position of the chargepad in the instances of no detection in some frames. We use SORT \cite{SORT} for online and real-time tracking of the detected chargepad. Once the bounding box is detected and tracked, the vehicle's angle towards the detected chargepad is estimated based on our previous work \cite{DeepTraierAssist}.\par

\begin{figure}[t]
\centering
\captionsetup{singlelinecheck=false, font=small, belowskip=-4pt}
\includegraphics[width=\columnwidth]{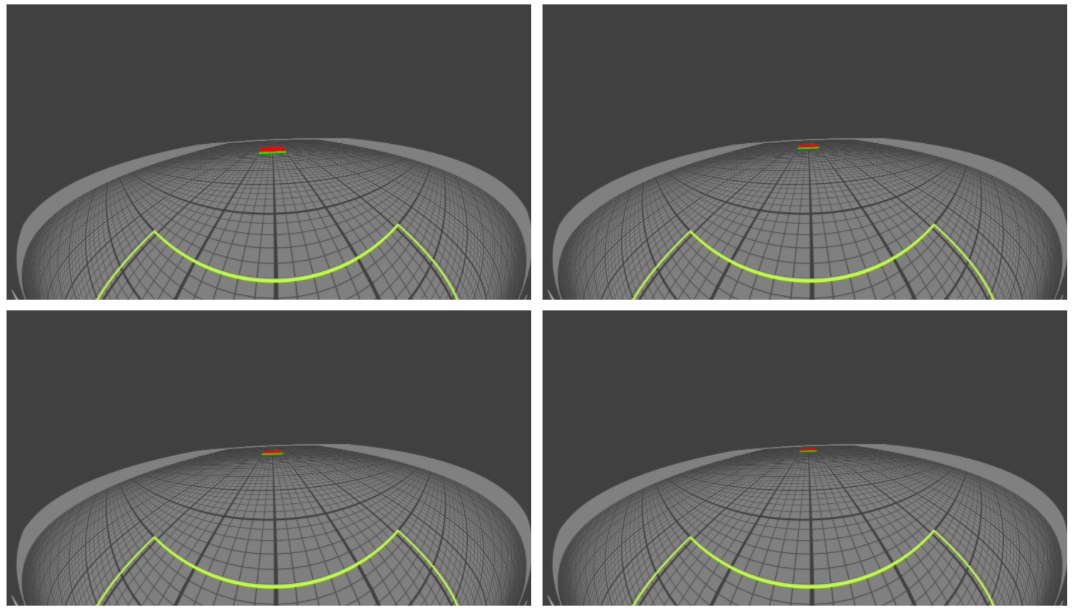}
\caption{\textbf{Synthetic analysis of the pixel size of chargepads at various distances from the vehicle.}}
\label{fig:range}
\vspace{-4mm}
\end{figure}

As the pattern looks relatively small when the vehicle is far, the detection will only start when the vehicle approaches a reasonable distance from the chargepad. To analyze this, we modeled what a typically sized chargepad (60cm~$\times$~60cm) looks like in a fisheye image, as shown in Figure \ref{fig:range}. The size in the image at various distances is given in Table \ref{tab:resolution}. The image height of the chargepad is only 12 pixels at 6m, which we expect to be the upper limit of detection (this is analyzed later in \S\ref{sec:results}).\par
\begin{table}[t]
\centering
\captionsetup{singlelinecheck=false, font=small, belowskip=-4pt}
\caption{\textbf{Resolution of chargepad }at distances simulating a 2Mpix fisheye camera and a target of 900x800x70 mm.}
\begin{tabular}{@{}ccccc@{}}
\toprule
\textit{Distance (m)}  & 3 & 4 & 5 & 6 \\ \midrule
\textit{Size (pixels)} & $100 \times 31$ & $81 \times 23$ & $65 \times 20$ & $54 \times 12$  \\ 
\bottomrule
\end{tabular}
\label{tab:resolution}
\vspace{-0.2cm}
\end{table}
\subsection{Online Self-supervised  Learning}

The fundamental problem with the conventional offline trained machine learning approach is the inability of the computer vision network (the deep learning model) to identify the non-trained, or previously unseen, classes of objects at inference time. Specifically, for our proposed architecture, this means that it is impossible to cover all types of chargepad objects before deployment, due to the fact that every vehicle manufacturer will have different visual styles for their chargepads (Figure \ref{fig:diffCpads}). In a scene where a previously unseen chargepad design is present, there is a strong likelihood that any machine learning approach will fail to detect the chargepad.

\begin{figure}[t]
\centering
\captionsetup{singlelinecheck=false, font=small, belowskip=-4pt}
\includegraphics[width=\columnwidth]{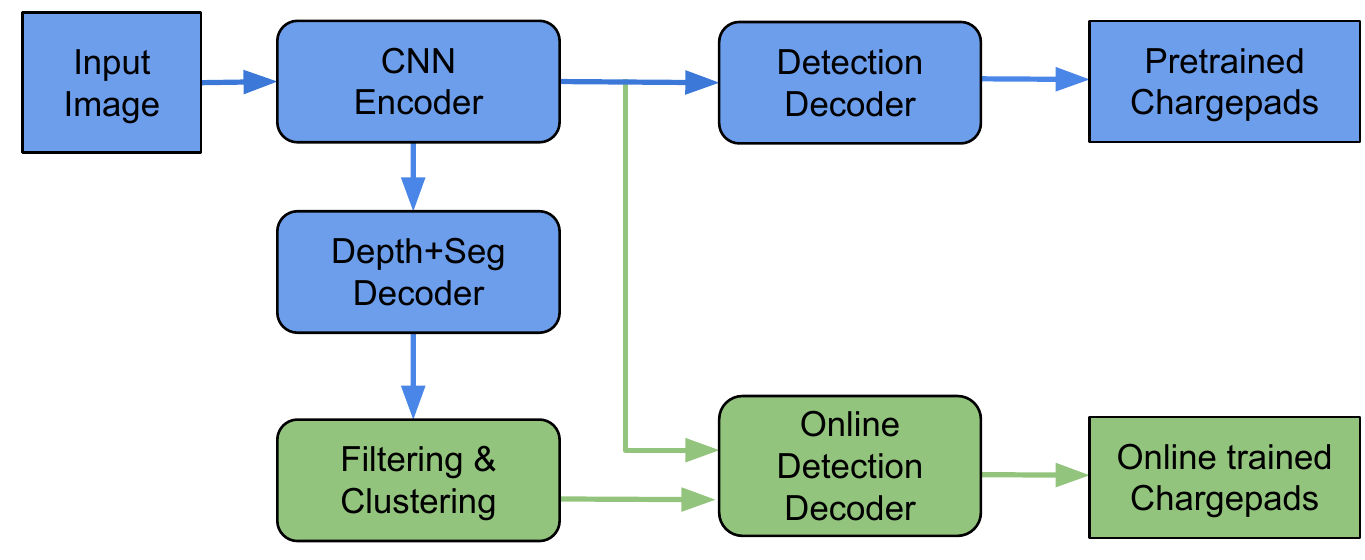}
\caption{\textbf{Illustration of online learning framework. Blue blocks denote the offline trained model.} Green blocks denote the online training model, which uses post-processing of depth and segmentation guiding by backtracking of the driver's final position of chargepad location.}
\label{fig:onlinelearning}
\vspace{-0.4cm}
\end{figure}

While extensive work has been done in unsupervised learning approaches in different domains \cite{scan, unsup}, supervised learning approaches outperform unsupervised approaches for visual recognition tasks \cite{Beyer2019}. However, it is not feasible to manually annotate the online data as it is generated during parking maneuvers. Doing so would incur a high cost, as this requires data to be transferred from the vehicle to a facility where manual annotation can occur, the network to be re-trained, and downloaded again to the vehicle. This makes the goal of online auto-annotation for use in self-supervised learning much more attractive.
Figure \ref{fig:onlinelearning} illustrates our proposed online learning framework. At inference time, the system detects the pre-trained class of objects, as usual, which can include an incomplete set of pre-trained chargepads. If the end-user attempts to engage the chargepad auto-alignment but fails to detect a previously unseen chargepad, the driver will have to manually complete the alignment. In this case, the alignment need not be perfect. For a single charging cycle, non-optimal charging performance will be acceptable. However, this manual alignment is used to provide the auto-annotation for online learning.\par
\begin{figure}[t]
\centering
\captionsetup{singlelinecheck=false, font=small, belowskip=-4pt}
\includegraphics[width=\columnwidth]{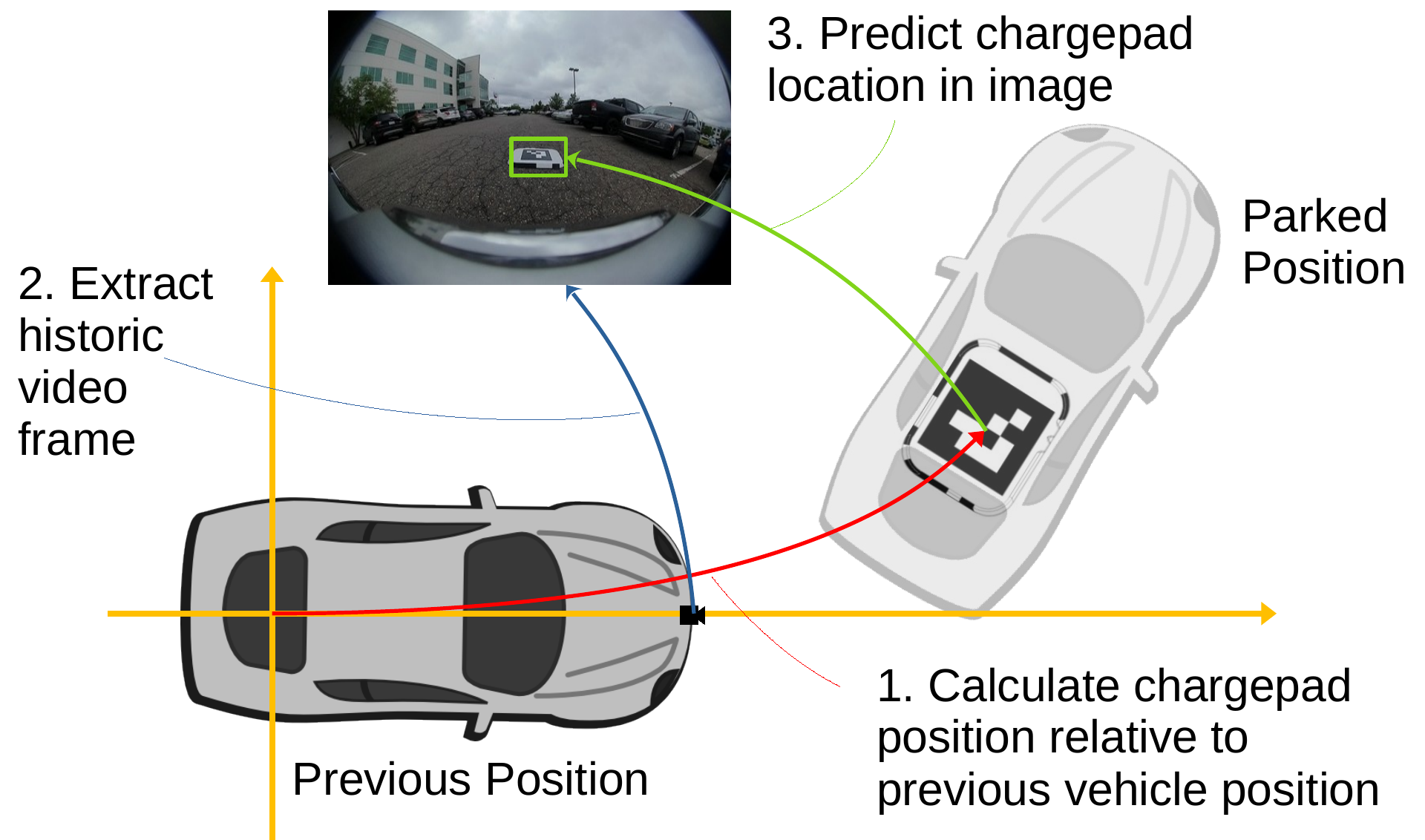}
\caption{\textbf{Employing Visual SLAM to predict the position of the chargepad in an image corresponding to a previous vehicle position.}}
\label{fig:predict_chargepad}
\vspace{-0.2cm}
\end{figure}

Thus, the approach for self-supervised learning of chargepads is as follows. During the manual parking maneuver, video from the cameras is recorded after a failed detection of a chargepad. Once the vehicle has been manually parked, the recorded video frames are then available to the online self-supervised training algorithm. Video frames are retrieved from the storage in sequence, and Visual SLAM and vehicle odometry \cite{Eising2022} is used to predict the position of the chargepad in each frame of recorded video (Figure \ref{fig:predict_chargepad}). However, due to issues with the accuracy of the manual alignment of the vehicle (as already discussed) and due to drift with odometry over time, this auto-annotation will have low accuracy. For that reason, we use road segmentation and depth outputs with uncertainty maps of the MTL network to refine the annotation ahead of training. Depth is used to find the ground plane to guide the object center to extract features and reason the depth from its contact point as a simple filter to remove any objects above the ground. The ground segmentation is used to refine the detection by filtering the ground surface, leaving just the chargepad as a refined detection. This refined auto-annotation of the chargepad objects can then be used to update the training of the MTL Object Decoder.\par
\subsection{Visual SLAM based Environment Localization}

Simultaneous localization and mapping (SLAM) is a technique for estimating sensor motion and reconstructing structure in an unknown environment. When a camera is (or multiple cameras are) used for this purpose, it is called Visual SLAM. The vision sensor can be a monocular, stereo vision, Omnidirectional (360-degree), or Red Green Blue Depth (RGBD) camera. Visual SLAM is composed of mainly three modules: initialization, tracking, and mapping. Both \cite{vslam} and \cite{Fraundorfer2012} provide extensive surveys of Visual SLAM techniques.

\begin{figure}[t]
\centering
\captionsetup{singlelinecheck=false, font=small, belowskip=-4pt}
\includegraphics[width=0.7\columnwidth]{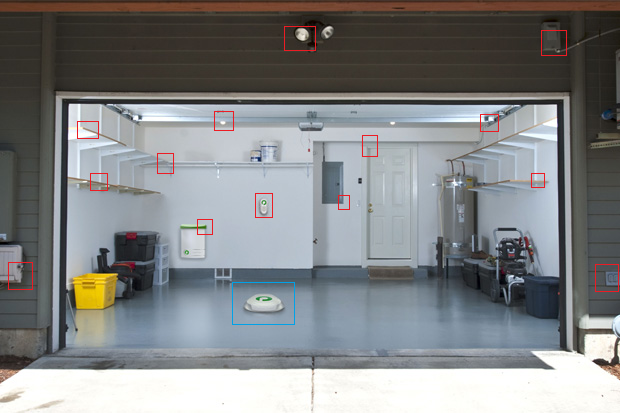}
\caption{\textbf{Examples of visual features in an image.} The exact features depend on the feature Extraction and Description algorithm used \cite{feat_etract_survey}.}
\label{fig:visual_features}
\vspace{-2mm}
\end{figure}

We base our Visual SLAM enhancement for chargepad alignment on previous work on trained parking using SLAM \cite{tripathi2020trained}. In the next section, we describe the overall Visual SLAM architecture employed in the proposed system. However, it is pertinent to provide some details here, though the reader is referred to \cite{tripathi2020trained} for complete details. The motivation is illustrated in Figure \ref{fig:visual_features} where feature locations (marked in red boxes) relative to the chargepad (marked in the blue box) are found. This will enable localization at a farther distance when the chargepad is poorly visible.\par
Note that at the feature detection stage, the location of the chargepad is not required to be known. When the bundle adjustment step of the SLAM is performed, a complete feature map is built of the trajectory. Subsequently, when the vehicle is aligned with the chargepad, the location of the vehicle and thus the chargepad is known, and the position of the chargepad can be stored relative to the feature map. This enables the future relocalization of the vehicle to the chargepad, even when the chargepad cannot be directly detected.

\begin{figure}[t]
\centering
\captionsetup{singlelinecheck=false, font=small, belowskip=-4pt}
\includegraphics[width=0.8\columnwidth]{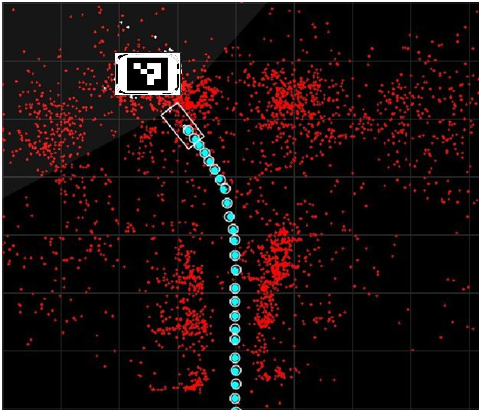}
\caption{\textbf{3D feature layout for a single trajectory.}}
\label{fig:slam1}
\vspace{-0.4cm}
\end{figure}

The trained trajectory following operates in two distinct phases: ``teach and repeat''. During the teaching phase, the driver manually drives to the chargepad location (within the range of the chargepad detector described above). Visual features are extracted from the video stream and matched between frames. Oriented FAST and rotated BRIEF (ORB) based SLAM \cite{MurArtal2015} has become almost standard in many applications. However, we have found that Accelerated KAZE (AKAZE) \cite{alcantarilla2011fast} is a more robust feature for the repeat phase, and as such, we use AKAZE for the feature extraction and matching. Feature matching is computationally expensive, however, \cite{tripathi2020trained} discuss an embedded Visual SLAM system for automated parking. The end goal of the SLAM approach is bundle-adjustment, that is, using Levenberg-Marquardt to iteratively refine the 3D location of the features and the position of the vehicle across the entire trajectory (e.g., Figure \ref{fig:slam1}). For a complete review of visual odometry techniques, the reader is referred to \cite{yousif2015}.

The appropriately trained trajectory is loaded in the repeat phase, and the recorded bundle-adjusted keyframe feature descriptors are scanned and matched to visual features extracted from the current frame. Levenberg-Marquardt is then used here again. However, this time, only the vehicle's position is solved (as the 3D structure is resolved during the teaching phase). 

\subsection{Overall Architecture}

\begin{figure*}[t]
\centering
\captionsetup{singlelinecheck=false, font=small, belowskip=-4pt}
\includegraphics[width=\textwidth]{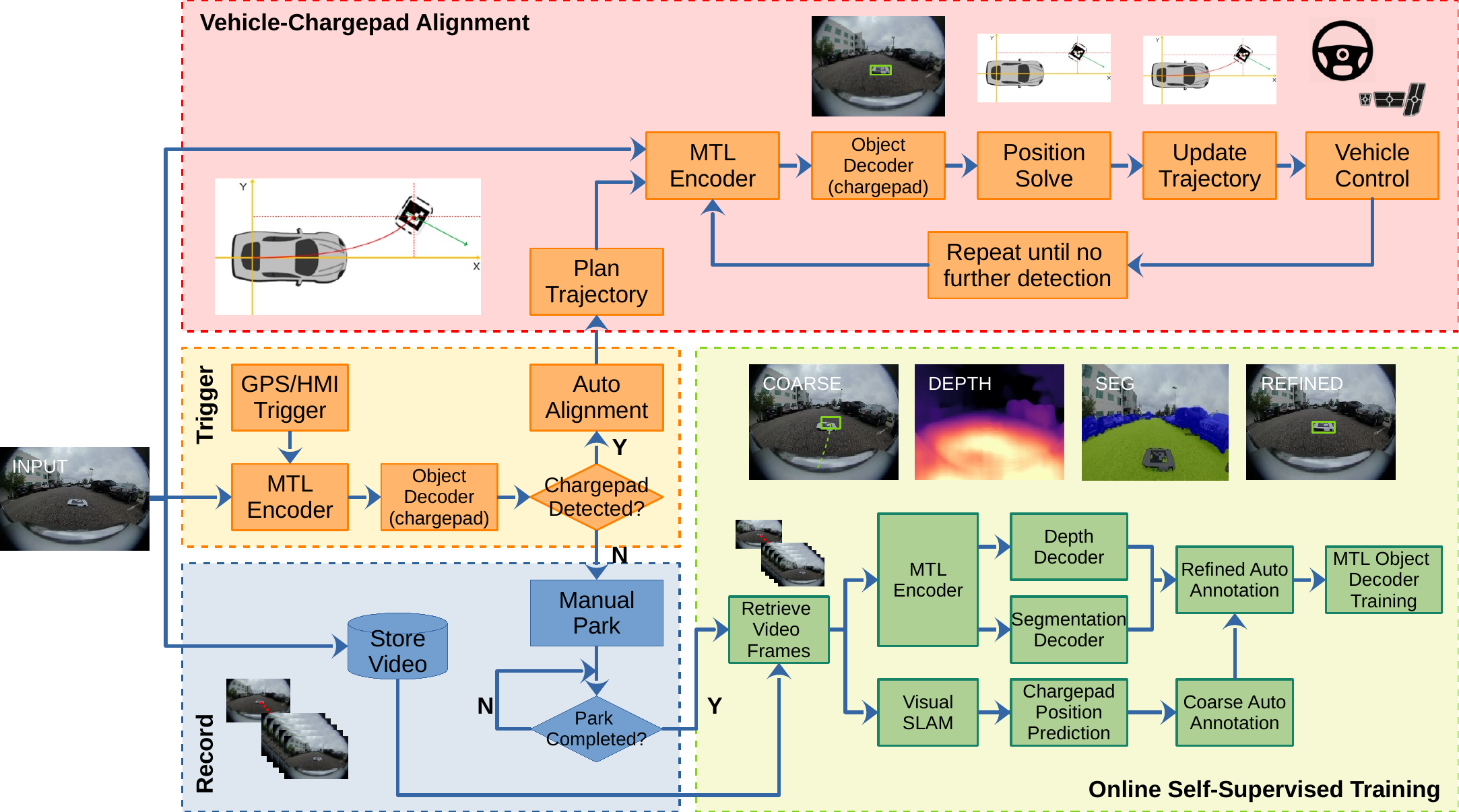}
\caption{\textbf{Overall system architecture for online chargepad learning and vehicle-chargepad alignment.}}
\label{fig:arch}
\vspace{-0.4cm}
\end{figure*}
In previous sections, we have discussed individual building blocks of the proposed approach. In this section, we will discuss the overall architecture of our proposed system, which is shown in Figure \ref{fig:arch}. We will explain each of the stages. Firstly, a Global Positioning System (GPS) or Human-Machine Interface (HMI) \textit{Trigger} prompts the system. For example, GPS can tell that the host vehicle is in the vicinity of the chargepad and thus triggers an attempt at detecting and aligning the vehicle with the chargepad. Alternatively, this trigger can be done via user input. If the pre-trained MTL detection network recognizes a chargepad (per Figure \ref{fig:onlinelearning}), then the host vehicle can proceed with the alignment. It should be noted that MTL Encoder is mentioned several times in Figure \ref{fig:arch}; however, it is just a single encoder that is running.\par

The mechanisms of vehicle alignment and control are not the subject of this paper; however, vehicle control has been a reasonably well-understood problem for several decades \cite{Fenton1970}, though there is more recent research in the topic \cite{ALCALA20181}. In Figure \ref{fig:arch}, we show a high level description of that the \textit{Vehicle-Chargepad Alignment} implementation would look like. Given an initial detection and localization of the chargepad, the trajectory that the vehicle should follow can be calculated. To account for drift, the trajectory can be iteratively updated based on further chargepad detections, which will account for any drift. Eventually, the chargepad will pass out of the view of the cameras (e.g., when it passes under the body of the vehicle). The vehicle will rely solely on odometry to guide it to the final position.\par

In the case that the pre-trained network fails to detect a chargepad (for example, if the chargepad we are attempting to detect is not of a type in the pre-trained network dataset), unfortunately, the user must manually align the vehicle with the pad. In this case, the system triggers the \textit{record} of the video scenes alongside the manual alignment of the vehicle with the chargepad. Thus, video frames from the point of engagement of the system through to park completion are available to the online training. This enables the {\em online self-supervised training approach} that is discussed previously.\par

An issue with the approach outlined thus far is that the range of the chargepad detector is limited to a max of 5 to 6m (depending on configuration). This can impact the ability of the vehicle to align itself with the chargepad without overly complex maneuvers. The range can be extended by increasing the cameras' resolution, but this has other impacts at a system level, not least the cost of the system. A Visual SLAM-based relocalization pipeline can be used in this case, as discussed earlier. We adapt the general approach of relocalization against a pre-trained trajectory \cite{tripathi2020trained} for building a relocalization map relative to the chargepad. We briefly summarize the algorithm here and refer to \cite{tripathi2020trained} for more details of the algorithm. A trajectory is learned by the extraction of visual features and map and pose optimization. The trajectory, including visual feature descriptors, is stored. In the repeat phase, features extracted from the live video stream are matched to the stored features. This matching is then used to solve the position of the vehicle against the stored trajectory. The role of path planning and vehicle control is to guide the vehicle to join and follow the previously-stored trajectory. \par

In this case, we can deploy the Visual SLAM-based relocalization to extend the range of the chargepad alignment. When the chargepad alignment functionality is triggered, we first search for the chargepad itself using the MTL output. In the case that this fails, the system will then attempt to match the pre-trained trajectory. The vehicle can then follow the pre-trained trajectory until the chargepad comes within range for direct detection and final alignment.\par

\subsection{Discussion}

It is worth taking a few moments to discuss some elements of the methodology that is proposed. At first glance, the overall system described by Figure \ref{fig:arch} would seem complex. However, it is important to realize that only certain parts of the system run at any given time. When the chargepad alignment functionality is triggered, an attempt is made to detect the chargepad in the vicinity. In the case that this is successful, the \textit{Vehicle-Chargepad Alignment} part of the system is executed, which is a standard paradigm for vehicle automation. In the case that the chargepad detection fails, the \textit{Record} branch is executed as the driver manually aligns the vehicle with the chargepad. Once the parking is completed, the \textit{Online Self-Supervised Training} part of the system is executed, whereby the MTL component is re-trained using the auto-annotated video samples of the previously unseen chargepad. Thus, the different parts of the overall system are largely decoupled from one another, reducing the overall complexity.

This also points to the feasibility of implementation in an embedded system, which is a very real concern in any automotive system. None of the overall components of the system run in parallel. Each of the \textit{Record}, \textit{Online Self-Supervised Training}, and \textit{Vehicle-Chargepad Alignment} parts of the architecture have exclusive access to the computational resources. The \textit{Online Self-Supervised Training} is the most computationally intensive part of the system. However, this part does not need to run in real-time. It runs based on previously recorded data and only when the vehicle is parked and charging. Naturally, one still does not want this part of the system to take more than a few minutes to complete the online training, and as such significant computer resources are still required, for example, in the Nvidia Drive Series of embedded platforms for self-driving cars.

It is also natural to consider which visual feature is best for SLAM-based relocalization that is discussed previously. Table \ref{tab:featcompare} presents results of comparison of various feature detectors and descriptors on the dataset described in \cite{tripathi2020trained}. The implementation is taken from OpenCV 3.0 \cite{kaehler2016learning} and we use the corresponding acronyms. Different metrics of our use case namely relocalization success rate, embedded run-time on an ARM A57, accumulated position and orientation offset are tabulated. AKAZE \cite{alcantarilla2011fast} provides the best relocalization rate with reasonable run-time. Additionally, it should be noted that the ``teach and repeat'' Visual SLAM approach to extending the range of the chargepad alignment is limited to places where a user may regularly visit and where it would be feasible for the teaching phase to have been completed, for example their home or workplace. Outside of these areas, for example at a public charging facility, the use must rely solely on the direct detection of the chargepad, with the associated reduction in alignment range.
\begin{table*}[t]
\centering
\captionsetup{singlelinecheck=false, font=small, belowskip=-4pt}
\caption{\textbf{Comparison of various feature detectors and descriptors for our use case.} Detector and descriptor acronyms used here are as per OpenCV 3.0 \cite{kaehler2016learning}.}
\begin{tabular}{l|l|l|l|l}
\toprule
\textit{Detector\_Descriptor} & \textit{\begin{tabular}[c]{@{}l@{}}Relocalization \\ Success rate (\%)\end{tabular}} & \textit{Runtime (ms)} & \textit{\begin{tabular}[c]{@{}l@{}}Accumulated \\ Position Offset (m)\end{tabular}} & \textit{\begin{tabular}[c]{@{}l@{}}Accumulated  \\ Orientation Offset (degree)\end{tabular}} \\ \hline
AKAZE\_KAZE\_UR & 58.21 & 97  & 0.15 &  2.74 \\
GFTT\_ORB       & 49.23 & 39  & 0.17 & -0.97 \\
AKAZE\_KAZE     & 49.99 & 102 & 0.11 & 2.96  \\
FAST\_ORB       & 54.66 & 35  & 0.28 & 3.33  \\
GFTT\_DAISY     & 52.35 & 171 & 0.13 & 2.88  \\
FAST\_BRISK     & 45.44 & 41  & 0.16 & 3.45  \\
STAR\_ORB       & 36.31 & 42  & 0.12 & -0.03 \\
GFTT\_BRISK     & 45.58 & 50  & 0.13 & 3.38  \\
FAST\_DAISY     & 54.74 & 156 & 0.17 & 4.2   \\
GFTT\_FREAK     & 39.13 & 41  & 0.17 & 1.3   \\
STAR\_FREAK     & 7.72  & 45  & 0.08 & 0.07  \\
STAR\_DAISY     & 33.45 & 160 & 0.13 & 1.26  \\
ORB\_ORB        & 42.35 & 48  & 0.22 & 2.79  \\
\toprule
\end{tabular}
\label{tab:featcompare}
\end{table*}

\section{Results}
\label{sec:results}

\subsection{Dataset and hardware details}

The chargepad training dataset comprises 375 sequences with roughly equal indoor and outdoor scenes, all containing AruCo patterns on the chargepad. The test sequences were independently created from similar environments. Synthetic images were created with the chargepad to enable evaluation of additional road or pavement textures. The test set consists of 200 sequences with a split of 65 indoor, 64 outdoor, and 71 synthetic scenes.
The dimensions of the chargepad used are $760mm \times 620mm \times 60mm$. The images were both captured from front and rear cameras. This is because chargepad detection and tracking can be in both forward and reverse directions.  The evaluation of synthetic images was purely based on only real image-based training to understand generalization to new road textures.
In order to test a different pattern, we have augmented our AruCo dataset. Given the calibration of the cameras, the annotation of the chargepad in the images, and the physical dimensions of the chargepad, we could overlay the Valeo logo on the images with perspective added. The orientation of the chargepad is not known, and as such this part of the augmentation was manual. Due to the manual intervention required, we only augmented 50 sequences in this manner. The colour of the Valeo logo is matched to the pixel values of the chargepad top.

For the baseline training, we train the model for 100 epochs on a 24GB Titan RTX with a batch size of 24. We use a learning rate of ${4 \times {10}^{-4}}$, which we reduce to ${{10}^{-5}}$ after 85 epochs for another 15 epochs.  We use Pytorch to implement the models and the Ranger (RAdam~\cite{liu2019radam} + LookAhead~\cite{zhang2019lookahead}) optimizer.
For the online training, the pre-trained MTL network~\cite{kumar2021omnidet} (Figure \ref{fig:perception}) has an additional chargepad decoder branch. 
The chargepad decoder has the same architecture as the object detection decoder from MTL. 
We initialize the chargepad decoder using the MTL object detection decoder weights. We only fine tune this chargepad decoder during online training for two reasons. Firstly, it is simpler to train a small part as the compute available on the vehicle is smaller than a typical more powerful training GPU. Secondly, we need to maintain the other MTL outputs for the generic automated driving pipeline. 

As the online training data is very small, we use extensive data augmentation techniques to increase the dataset size by a factor of ~30X using  Albumentations \cite{buslaev2020albumentations} tool. We apply photometric augmentations including change of brightness, gamma, contrast \& color and geometric augmentations like rotations, affine, perspective \& elastic transformations and other modifications like Gaussian blur and addition of noise. The augmentation step was crucial to facilitate a stable online training. When the online learned system is performing inference, the detection accuracy can be indirectly inferred by efficiency of charging system or lack of charging.

\subsection{Evaluation}

\begin{figure}[t]
\captionsetup{singlelinecheck=false, font=small, belowskip=-4pt}
\centering
\includegraphics[width=\columnwidth]{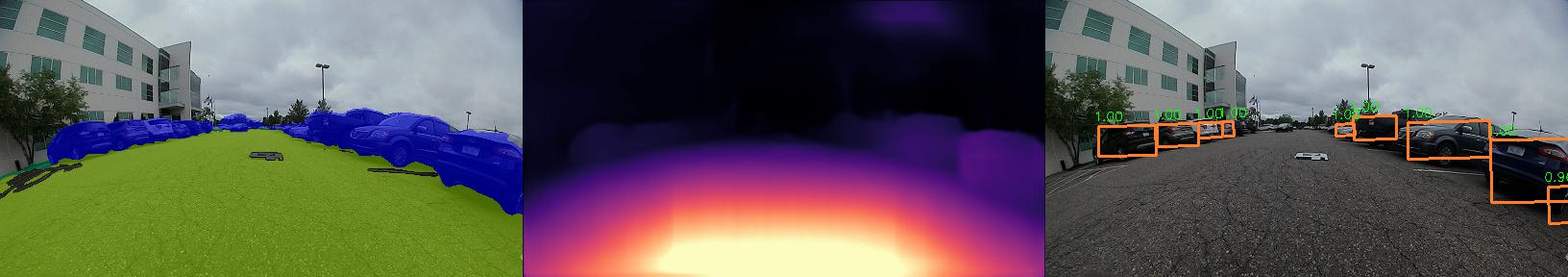}
\includegraphics[width=\columnwidth]{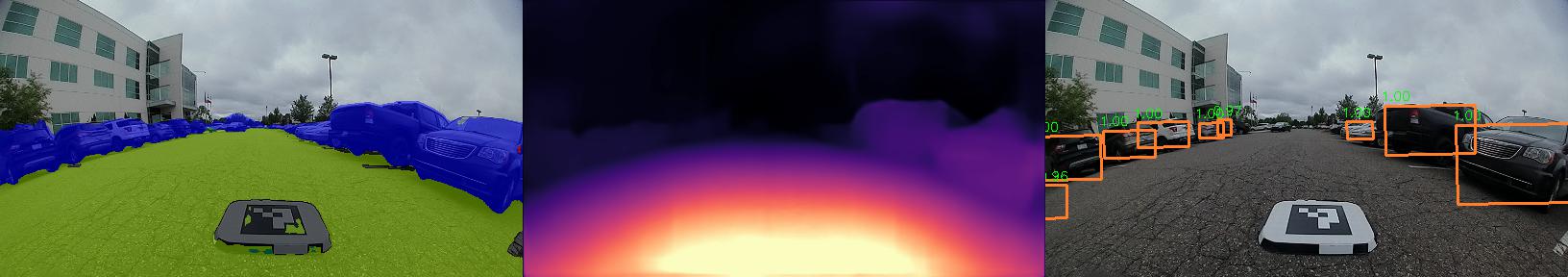}
\caption{\textbf{Detection of chargepad using depth and segmentation.}}
\label{fig:segdepth}
\vspace{-0.2cm}
\end{figure}

\begin{figure*}[t]
\captionsetup{singlelinecheck=false, font=small, belowskip=0pt}
\centering
\includegraphics[width=0.25\textwidth]{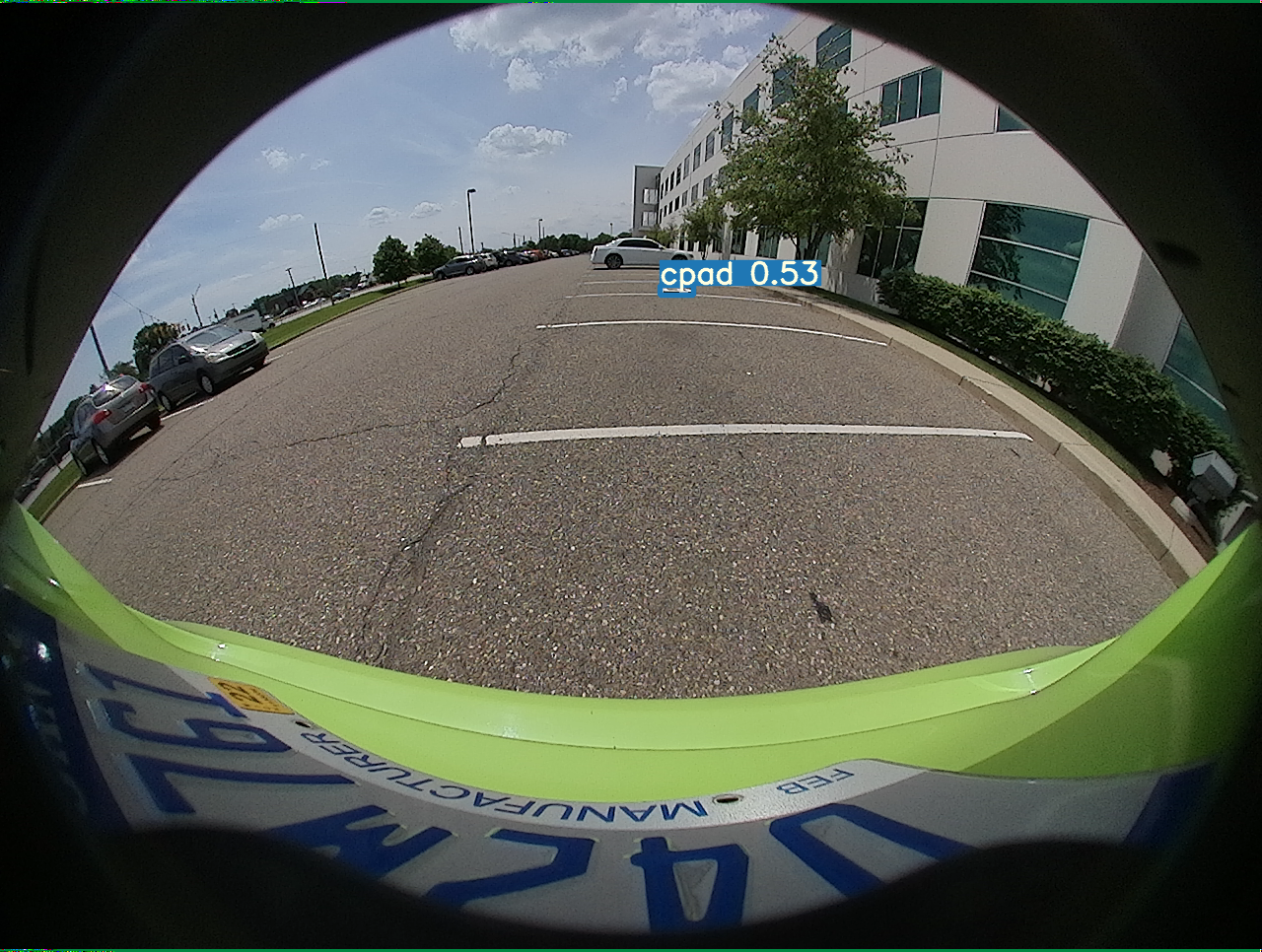}
\includegraphics[width=0.25\textwidth]{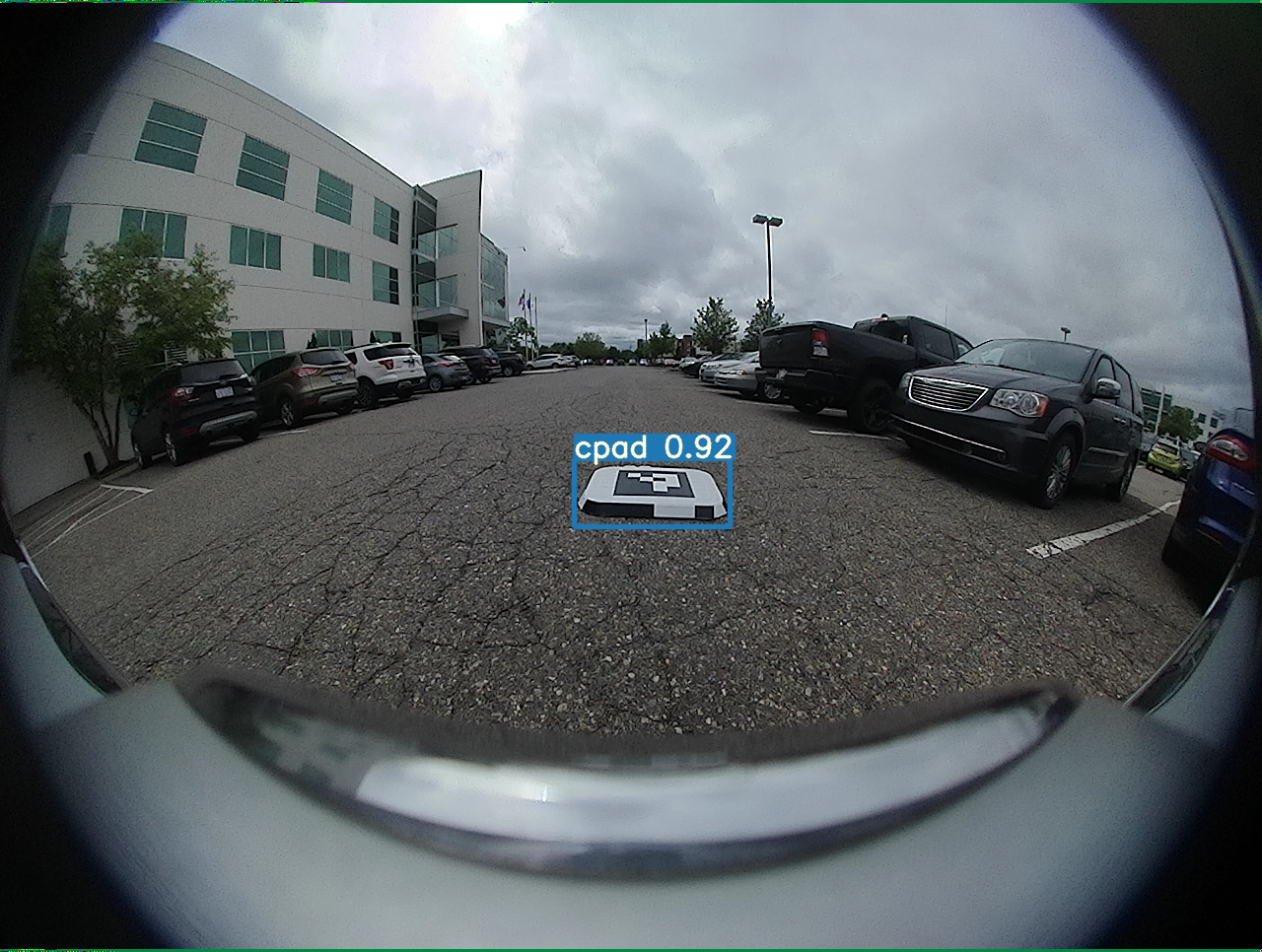}
\includegraphics[width=0.25\textwidth]{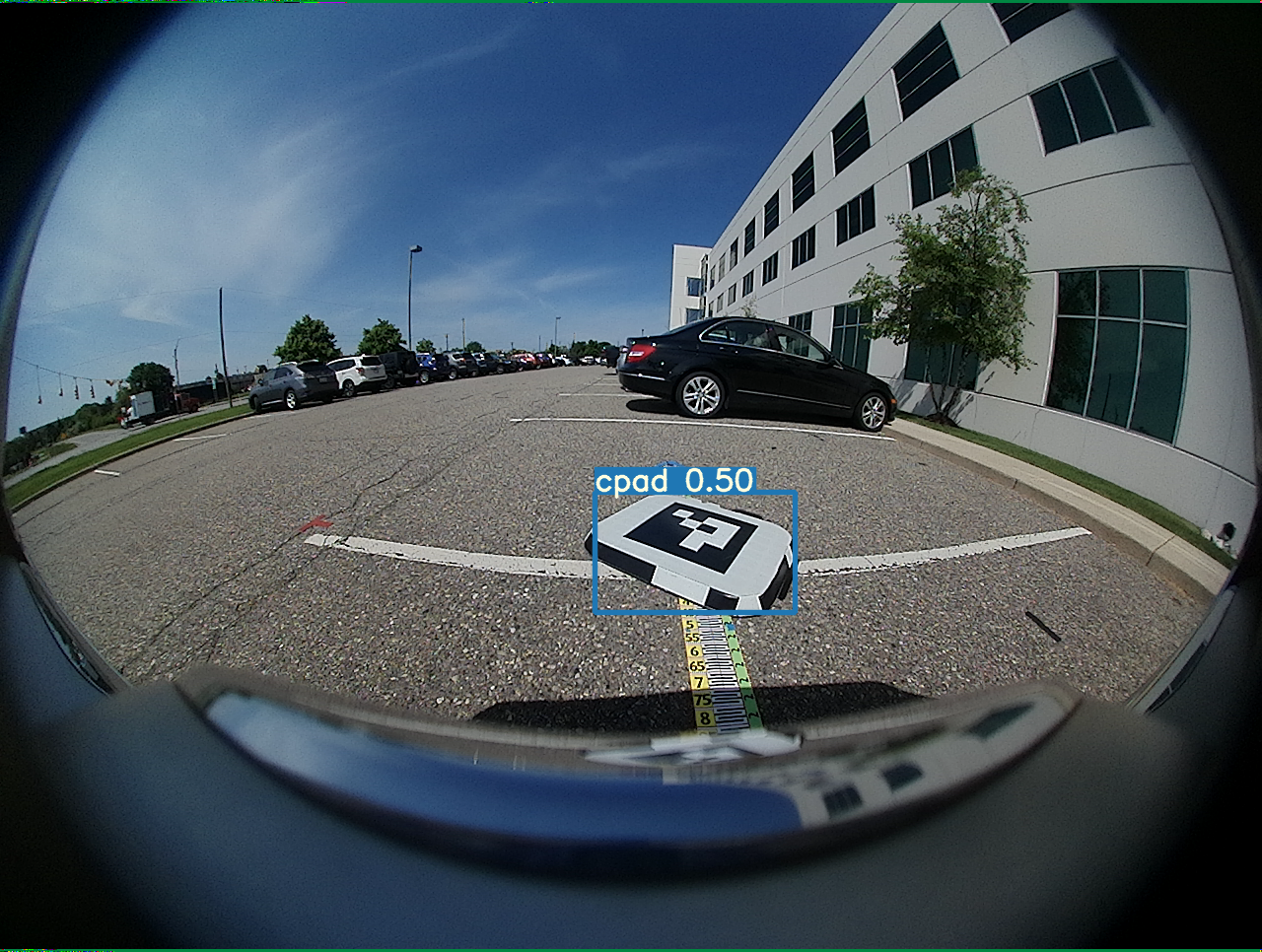}
\includegraphics[width=0.25\textwidth]{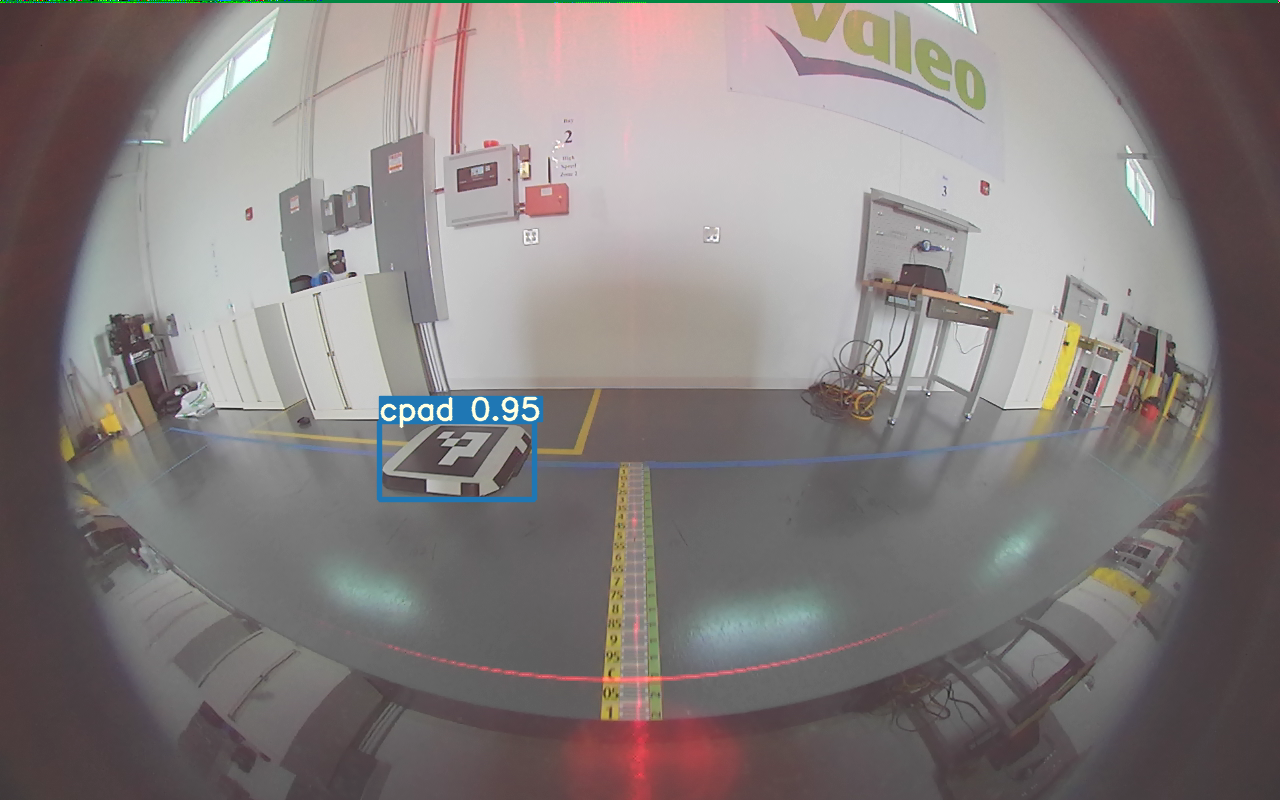}
\includegraphics[width=0.25\textwidth]{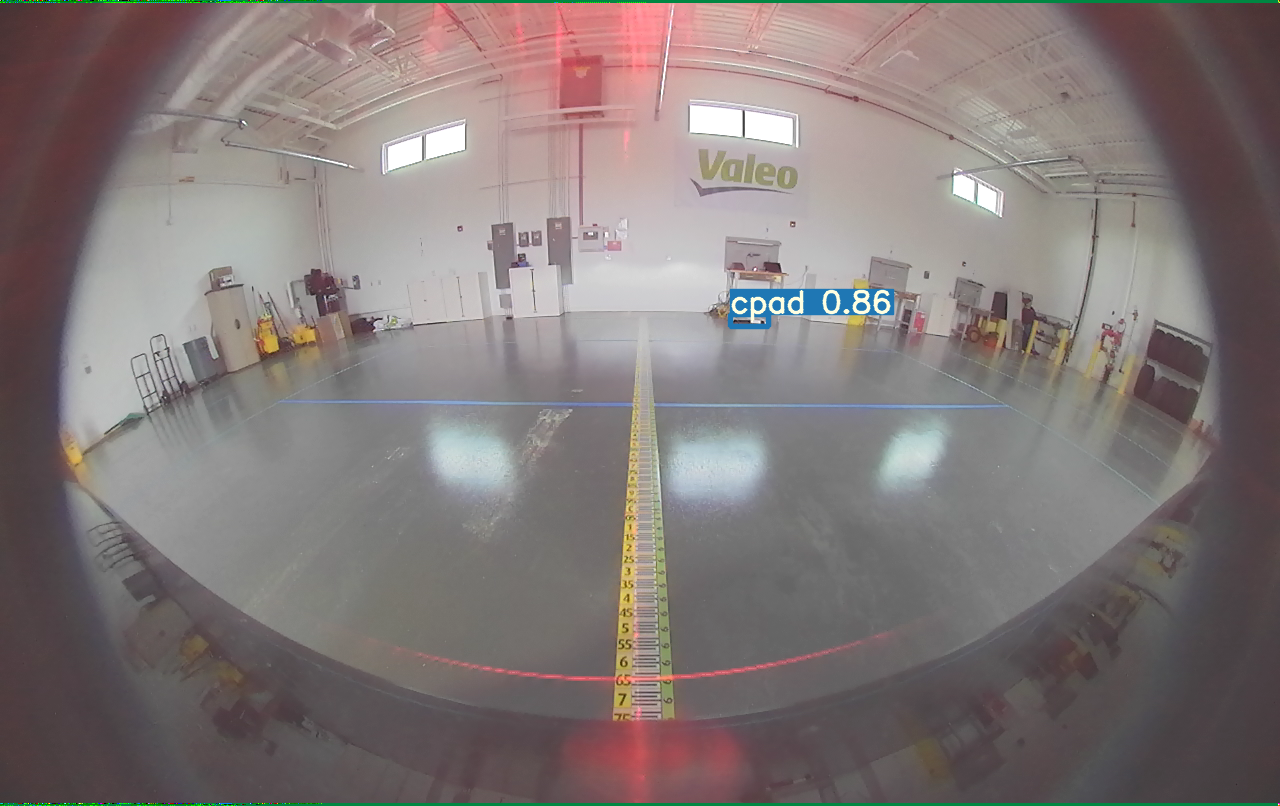}
\includegraphics[width=0.25\textwidth]{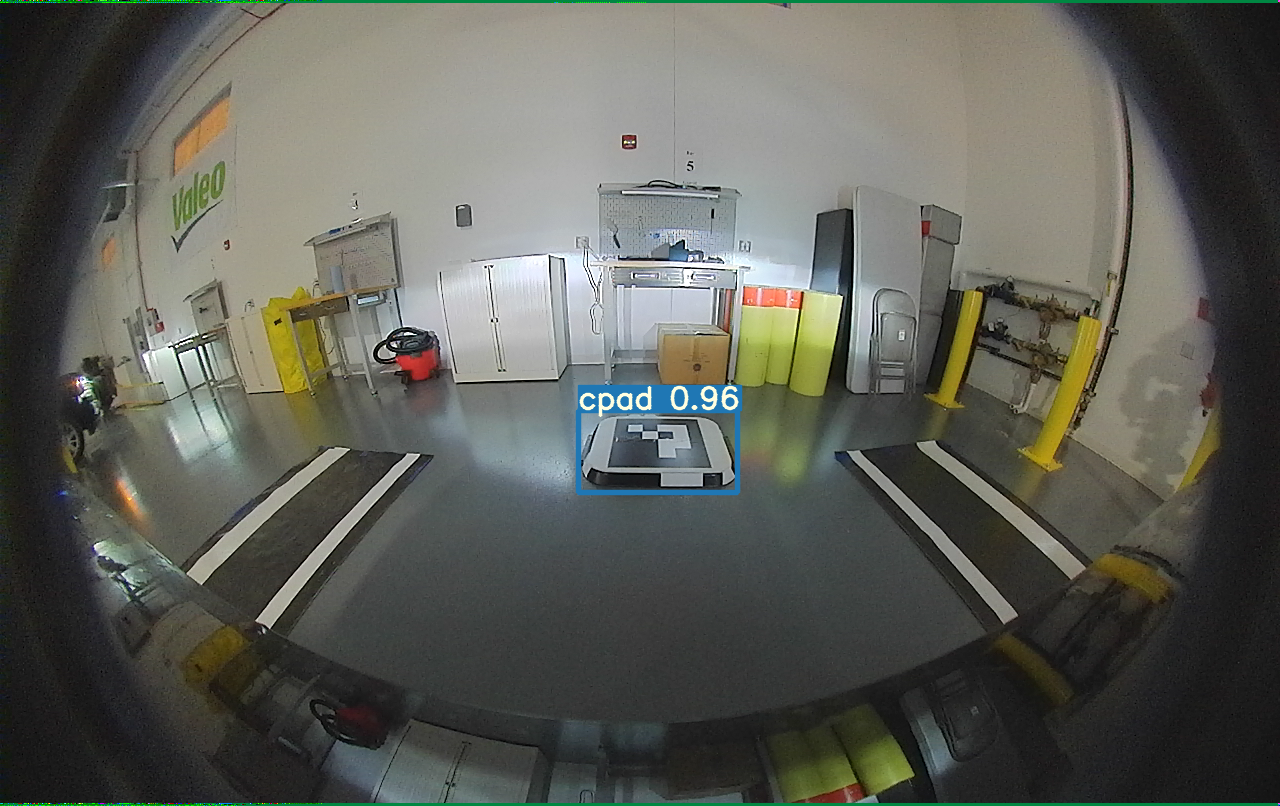}
\includegraphics[width=0.25\textwidth]{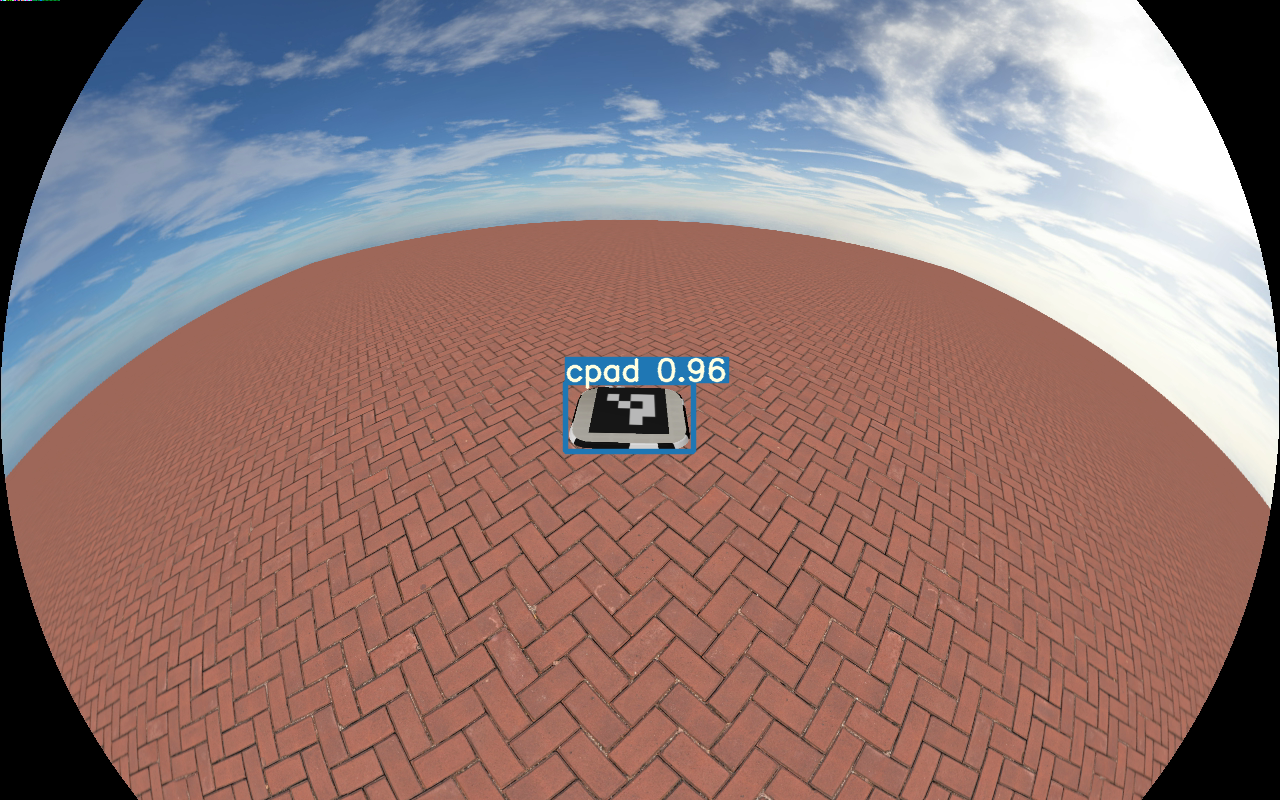}
\includegraphics[width=0.25\textwidth]{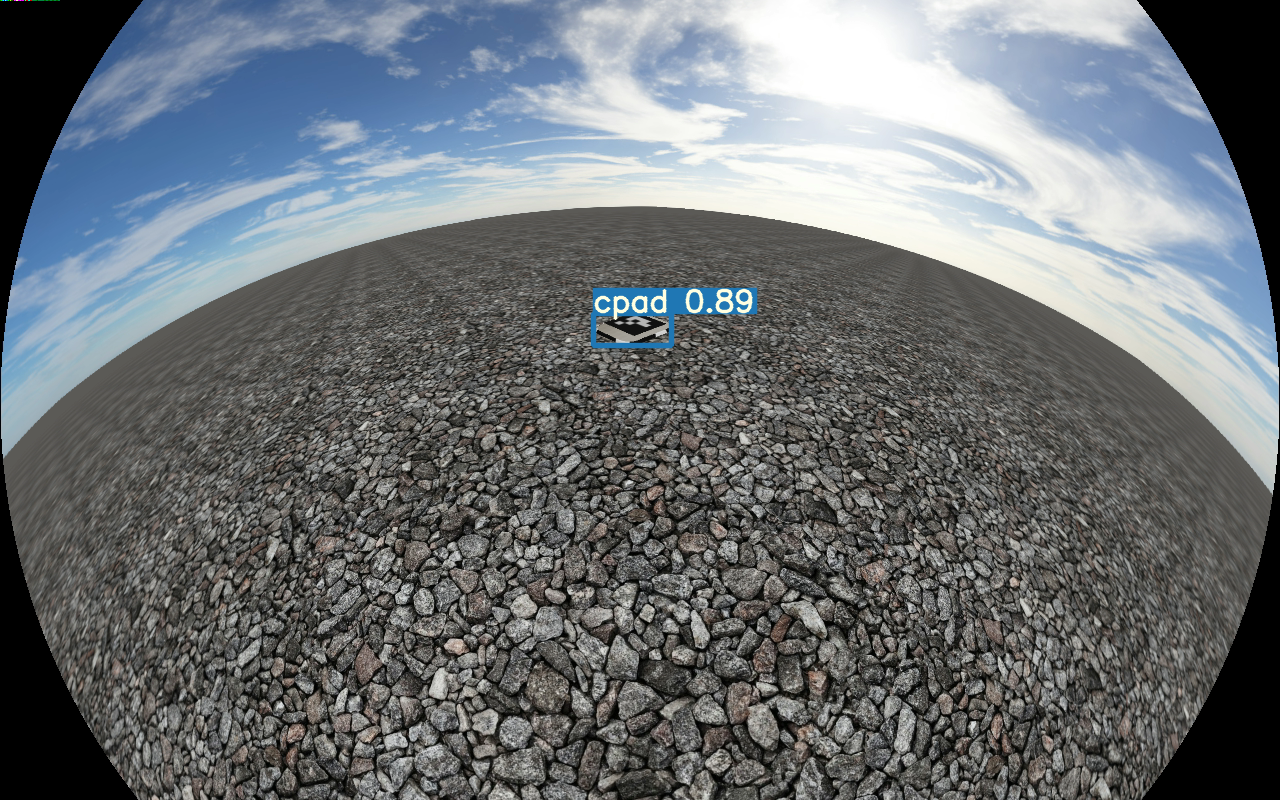}
\includegraphics[width=0.25\textwidth]{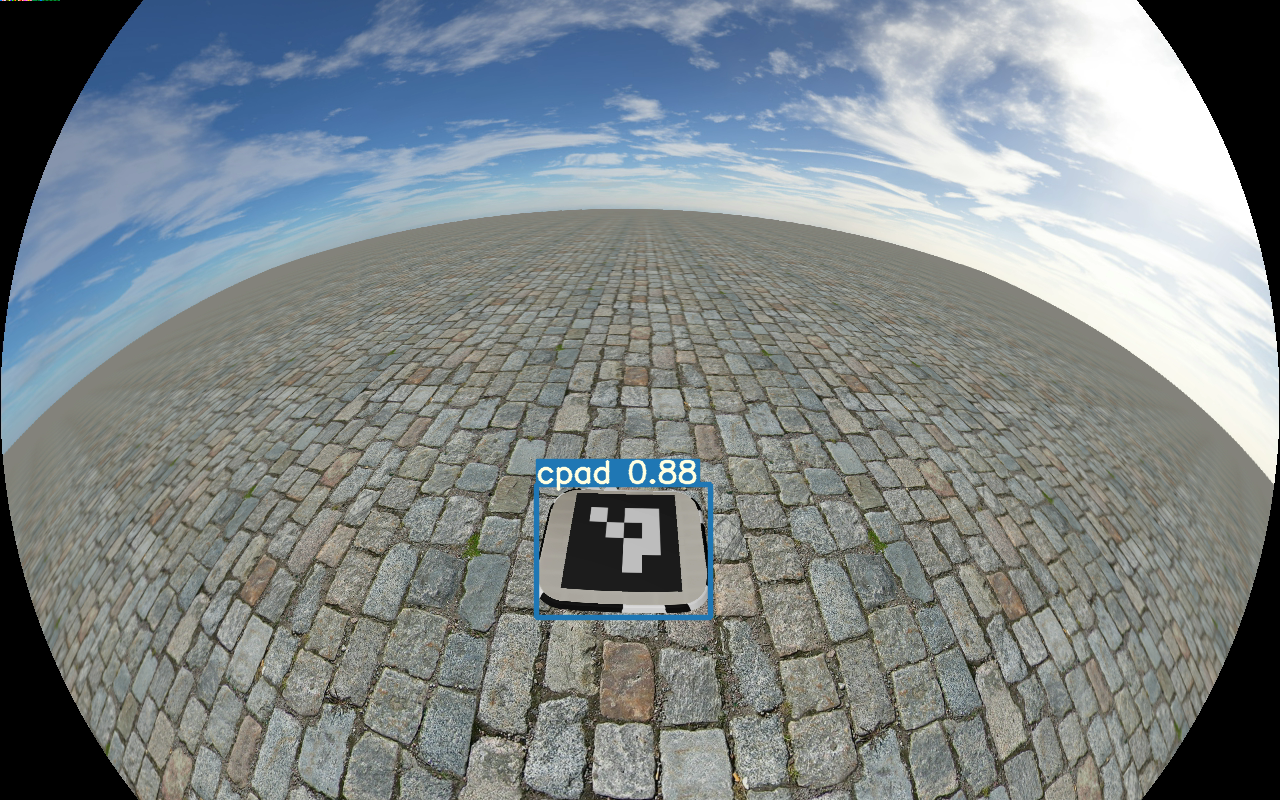}
\includegraphics[width=0.25\textwidth]{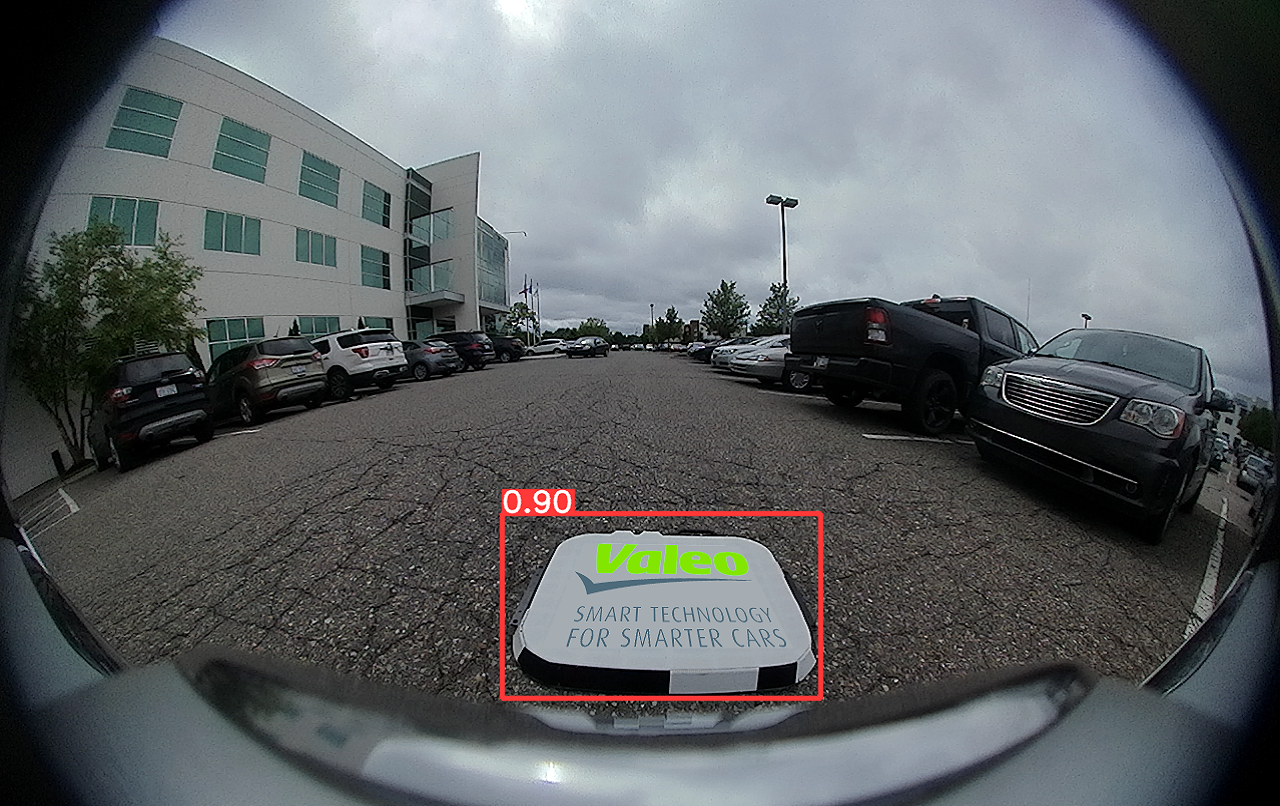}
\includegraphics[width=0.25\textwidth]{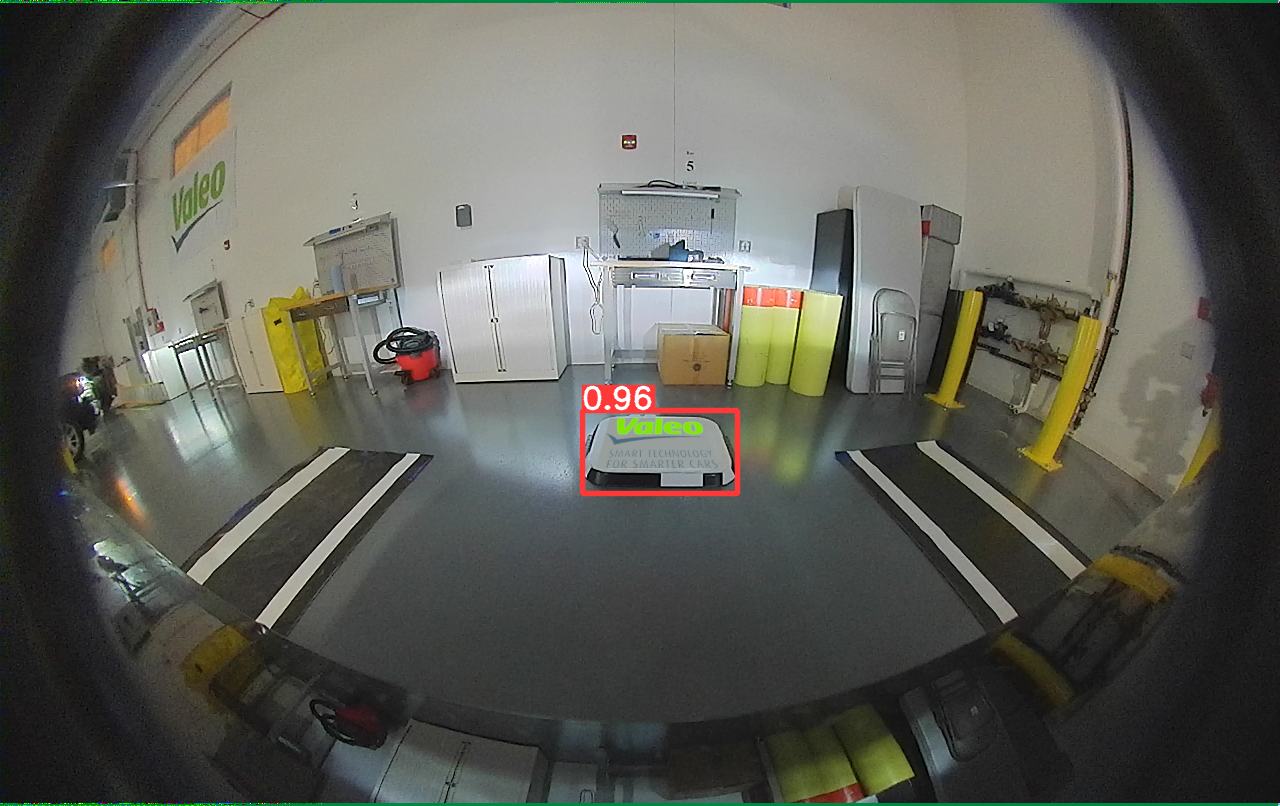}
\includegraphics[width=0.25\textwidth]{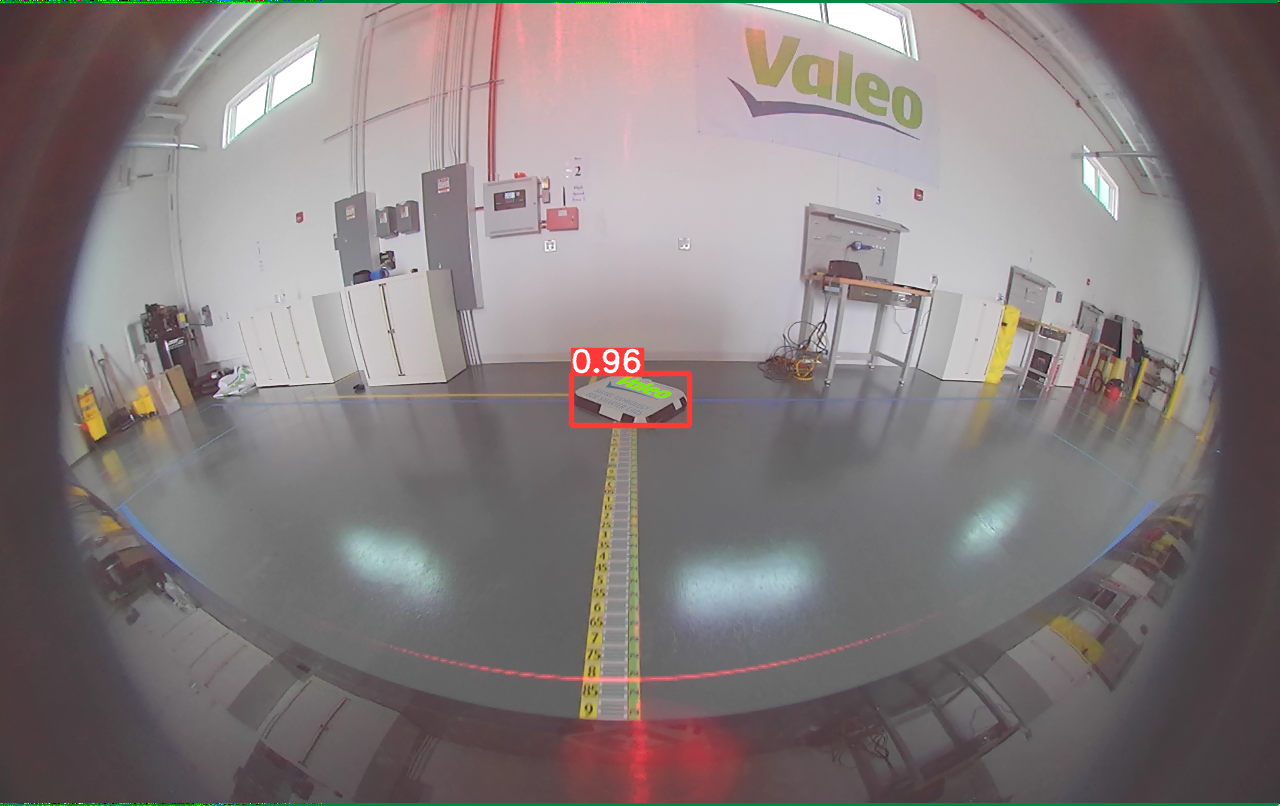}
\caption{\textbf{Qualitative results of chargepad Detection and Tracking in different scenarios namely outdoor (top), indoor (2nd row), synthetic (3rd row), and augmented Valeo logo (bottom).}}
\label{fig:detection}
\vspace{-0.4cm}
\end{figure*}

Firstly, Figure \ref{fig:segdepth} shows the multi-task baseline results illustrating the performance of segmentation and depth estimation of the chargepad and its surroundings. Figure \ref{fig:detection} illustrates the results of chargepad detection using the proposed method for outdoor, indoor, and synthetic scenes. The bounding box represents the detected chargepad on the image.
We use a confidence threshold of 25\% below which the detection is unreliable with false positives on other objects, such as vehicles.
Figure \ref{fig:detection} also illustrates reliable detection of chargepads at different ranges. To understand performance on various ground surface types, we also tested with a set of synthetic images (see the last row of Figure \ref{fig:detection}). The algorithm generalizes well from real to synthetic images.\par

Table \ref{tab:comparison} provides quantitative results of the detection performance using the average precision score on the test dataset with a 0.8 IoU threshold. We present a baseline offline learning algorithm, and the single sequence online learning using both the AruCo and the Valeo logo datasets.
We also tested a simple OpenCV based ArUco detector, but it achieves poor performance, particularly for ranges beyond 3 meters. Table \ref{tab:ablation_results} illustrates an incremental improvement of each feature presented in our system. It can be seen that the combination of backtracking using Visual SLAM and outlier rejection (Refined auto annotation) described previously using a combination of segmentation and depth leads to the best performance.\par
\begin{figure}[t]
\centering
\captionsetup{singlelinecheck=false, font=small, belowskip=-4pt}
\includegraphics[width=0.7\columnwidth]{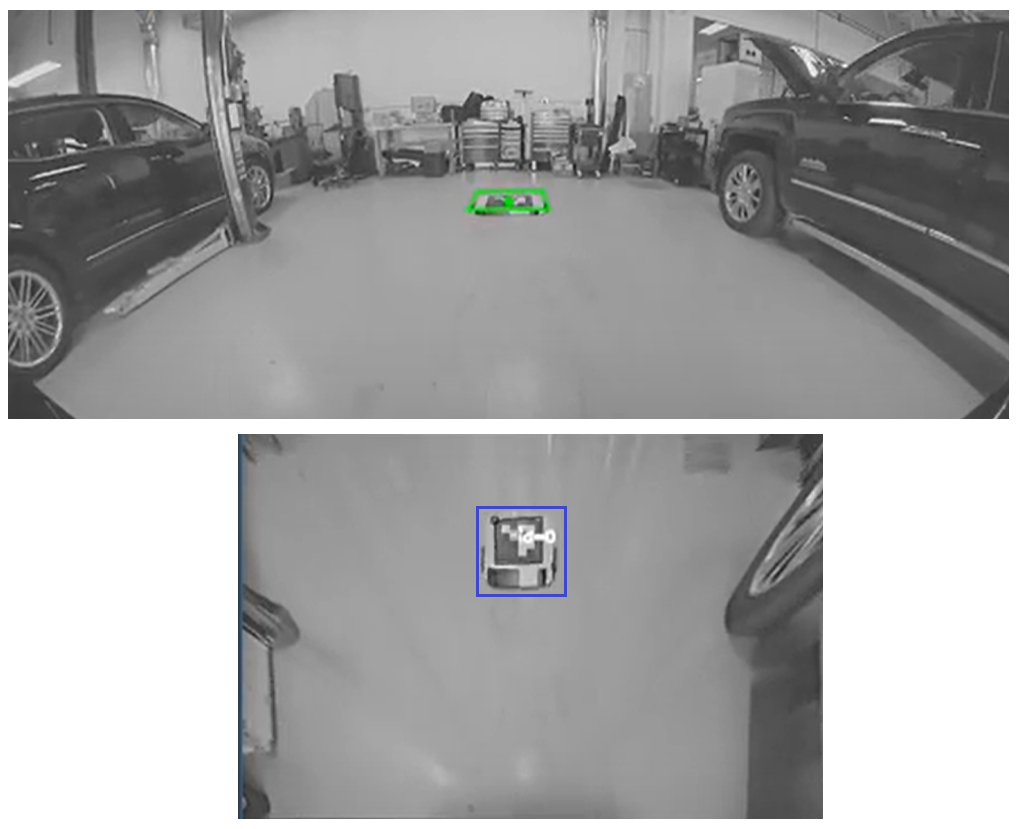}
\caption{\textbf{Top: Detection on regular fish-eye image view. Bottom: Detection on top view corrected image.}}
\label{fig:TopViewResults}
\end{figure}

\begin{table}[t]
\centering
\captionsetup{singlelinecheck=false, font=small, belowskip=-4pt}
\caption{\textbf{Comparative study of different approaches.}}
\begin{tabular}{@{}l|ccc@{}}
\toprule
\multicolumn{1}{c}{\textit{\textbf{Method}}} 
& \multicolumn{3}{c}{\textit{\textbf{Average Precision (IoU 0.80)}}} \\ 
\midrule
& \multicolumn{1}{l}{Indoor} 
& \multicolumn{1}{l}{Outdoor} 
& \multicolumn{1}{l}{Synthetic} \\
Baseline offline learning  & 93.5 & 98.9 & 95.3 \\
OpenCV ArUco detector      & 58.8 & 65.7 & 62.3 \\ 
Online learning - ArUco symbol & 85.5 & 90.3 & 87.9 \\
Online learning - Valeo symbol & 83.2 & 87.7 & -- \\
\bottomrule
\end{tabular}
\label{tab:comparison}
\vspace{-5mm}
\end{table}
\begin{table}[t]
\centering
\captionsetup{singlelinecheck=false, font=small, belowskip=-4pt}
\caption{\textbf{Ablation study of self-supervised online learning.}}
\begin{tabular}{@{}lc@{}}
\toprule
\textit{Method}   & \textit{Accuracy (IoU)} \\ 
\midrule
Segmentation Filtering Only   &         72.2    \\
\hspace{0.5cm} + Depth Filtering   &         75.3    \\
\hspace{0.5cm} + Odometry Backtracking  &    81.8    \\
\hspace{0.5cm} + Outlier Rejection &         85.4    \\
\hspace{0.5cm} + Visual SLAM       &         88.4    \\
\bottomrule
\end{tabular}
\label{tab:ablation_results}
\end{table}
For the previous experiments, we used the raw fisheye images without any undistortion. We explored normalization using top-view correction, mainly to improve the detection range. In Figure \ref{fig:TopViewResults}, we show a comparison of detecting the chargepad on raw fisheye and top-view. Top-view image improves the detection range significantly, as shown in Table \ref{tab:range}. However, it inhibits the usage of a shared encoder of the multi-task visual perception model. Thus, we also explore a top-view spatial transformer inspired by OFTNet \cite{roddick2018orthographic}. We insert a spatial transformer block in between the encoder and the YOLOv3 \cite{rashedfisheyeyolo} decoder. It slightly improves the detection range compared to the top-view corrected image. Finally, we show that combining Visual SLAM and CNN-based chargepad detection significantly increases the range compared to all other methods without requiring a top-view image transformation.
\begin{table}[t]
\centering
\captionsetup{singlelinecheck=false, font=small, belowskip=-4pt}
\caption{\textbf{Range study of different input representations.}}
\begin{tabular}{@{}l|c@{}}
\toprule
\multicolumn{1}{c}{\textit{Input}} & \textit{\begin{tabular}[c]{@{}c@{}}Average \\ Range (m)\end{tabular}} \\ \midrule
Fisheye                      & 5.2  \\
Top-view corrected           & 6.8  \\
Top-view spatial transformer & 7.1  \\
Fisheye + Visual SLAM        & 12.3 \\ \bottomrule
\end{tabular}
\label{tab:range}
\end{table}
\begin{table}[t]
\centering
\captionsetup{singlelinecheck=false, font=small, belowskip=-4pt}
\caption{\textbf{Localization and alignment accuracy in terms of final positional and angular offset}.}
\begin{tabular}{@{}llcc@{}}
\toprule
\multicolumn{1}{c}{\textit{Type}} &
  \textit{Scene} &
  \textit{\begin{tabular}[c]{@{}c@{}}Positional \\ Offset (cm)\end{tabular}} &
  \textit{\begin{tabular}[c]{@{}c@{}}Angular \\ Offset (deg)\end{tabular}} \\ 
\midrule
\multirow{3}{*}{Indoor}  & Scene 1 & 4.7 & 4.3  \\
       & Scene 2 & 3.6 & 5.5  \\
       & Scene 3 & 3.0 & 5.1  \\
\midrule
\multirow{3}{*}{Outdoor} & Scene 4 & 10.8 & 8.7  \\
       & Scene 5 & 9.0 & 9.2  \\
       & Scene 6 & 8.9 & 10.1 \\
\bottomrule
\end{tabular}
\label{tab:localization}
\vspace{-0.2cm}
\end{table}

Getting the exact alignment of the vehicle with the chargepad is difficult, as no automated ground truth data was readily available. However, we manually evaluated the final alignment of the vehicle against the chargepad on a small number of scenes (three indoor and three outdoor scenes), as presented in Table \ref{tab:localization}. It can be seen that at least five of the six scenes tested, we are within the $\pm 10$cm alignment accuracy required for optimal charging \cite{BIRRELL2015721}, with the final scene being only just outside. The alignment accuracy is slightly higher for indoor scenes, as the lighting tends to be more controlled. Additionally, in indoor scenes, the ground surface tends to be more even, reducing odometry drift when the chargepad is under the vehicle and no longer detected.

\subsection{Discussion}

While the overall performance of the proposed online learning system does not achieve the accuracy of the baseline offline learning (achieving approximately 8\% lower accuracy), this is to be expected. The baseline offline learning would be considered gold standard, and assumes that a fully annotated dataset of the target chargepad design is available. As we have discussed, this is not feasible in reality. In contrast, the online learning method must deal with noise introduced by the auto-annotation (i.e. noise in the segmentation, depth and Visual SLAM). We have shown that by using a combination of segmentation, depth and Visual SLAM, we achieve the highest accuracy in the auto-annotation, with the segmentation and depth offering significant refinement over the initial Visual SLAM based backtracking. The overall range of the alignment is the distance at which the vehicle can successfully start the chargepad alignment maneuver. This is increased slightly if we use a top-view correction or a top-view spatial transformer in our network. However, a much more significant increase in range is achieved when we combine Visual SLAM and chargepad detection. Comparing the detection of the Valeo and the ArUco patterns, the ArUco performs moderately better, as it is a more distinct visual pattern. Overall, we see that the final alignment with the chargepad is typically within the accuracy required for optimal wireless charging.

\section{Conclusion}

In this paper, we explored the application of wireless charging assistance for electric vehicles using widely available surround-view camera systems. Accurate alignment of the vehicle with the inductive chargepad is known to be critical for efficient charging. However, while visual techniques for alignment make much sense, not least due to the proliferation of cameras on vehicles in general, unseen chargepad designs will cause failure. We have proposed an online learning architecture to be scalable to any chargepad, using the known vehicle trajectory, augmented by object detection, semantic segmentation, and depth estimation, to auto-annotate unseen chargepads. Thus, when confronted with a previously unseen chargepad, the driver should only manually park once. Due to range limitations with the detection, we have also proposed an enhancement to using a Visual SLAM localization method. Using this combination of approaches, we have demonstrated an effective solution.

We have demonstrated that the proposed system for chargepad alignment on previously unseen designs has high accuracy. The overall precision of our proposed online learning model approaches that of the baseline offline learning model (which can be considered gold standard). Furthermore, it is evident that the combination of segmentation, depth and Visual SLAM, as proposed, is advantageous during the auto-annotation process. The range of chargepad alignment is significantly increased when a combination of Visual SLAM and chargepad detection is used during the maneuver. Finally, while the results for final vehicle alignment are limited for practical reasons, what is presented demonstrates that adequate accuracy is generally achieved for optimal wireless charging performance. We will release a portion of the dataset to encourage further research in this area.\par

\bibliographystyle{IEEEtran}
\bibliography{egbib}

\end{document}